\documentclass[11pt]{article}

\usepackage{graphicx}%
\usepackage{multirow}%
\usepackage{amsmath,amssymb,amsfonts}%
\usepackage{amsthm}%
\usepackage{mathrsfs}%
\usepackage[title]{appendix}%
\usepackage{xcolor}%
\usepackage{textcomp}%
\usepackage{manyfoot}%
\usepackage{booktabs}%
\usepackage{algorithm}%
\usepackage{algorithmicx}%
\usepackage{algpseudocode}%
\usepackage{listings}%

\usepackage{ifthen,pifont,latexsym}
\usepackage[T1]{fontenc}
\usepackage{color}
\usepackage{times,avant}
\usepackage[round,authoryear]{natbib}
\usepackage{amsmath}
\usepackage{amsfonts}
\usepackage{amssymb}
\usepackage{setspace}
\usepackage{comment}
\theoremstyle{thmstyleone}%
\theoremstyle{thmstyletwo}%

\theoremstyle{thmstylethree}%

\DeclareMathOperator*{\argmax}{arg\,max}

\def\NGSs{next generation sequencers (NGSs)
  \gdef\NGS{NGS}
  \gdef\NGSs{NGSs}}
\def\NGS{next generation sequencer (NGS)
  \gdef\NGS{NGS}
  \gdef\NGSs{NGSs}}
\def\SNP{single nucleotide polymorphism (SNP)
  \gdef\SNP{SNP}
  \gdef\SNPs{SNPs}}
\def\SNPs{single nucleotide polymorphisms (SNPs)
  \gdef\SNP{SNP}
  \gdef\SNPs{SNPs}}  
\def\MDS{multidimensional scaling (MDS)
  \gdef\MDS{MDS}}
\def\NCBI{National Center for Biotechnology Information (NCBI)
    \gdef\NCBI{NCBI}}
\def\MLE{maximum likelihood estimator (MLE)
  \gdef\MLE{MLE}
  \gdef\MLEs{MLEs}}
\def\ECDFs{empirical cumulative distribution functions (ECDFs)
  \gdef\ECDF{ECDF}
  \gdef\ECDFs{ECDFs}}
\def\ROC{receiver operating characteristic (ROC)
    \gdef\ROC{ROC}}

\begin{document}

\title{Structure of Classifier Boundaries:\\Case Study for a Bayes Classifier}

\author{Alan F. Karr\thanks{Temple University, Department of Statistics, Operations, and Data Science and Fraunhofer USA Center Mid-Atlantic, alan.karr@temple.edu}, Zac Bowen\thanks{Fraunhofer USA Center Mid-Atlantic; zbowen@fraunhofer.org}, Adam A. Porter\thanks{University of Maryland, Department of Computer Science, and Fraunhofer USA Center Mid-Atlantic; aporter@cs.umd.edu}; Regina Ruane\thanks{University of Pennsylvania, Computational Social Science Laboratory; ruanej@wharton.upenn.edu}}

\maketitle

\section{Introduction}\label{sec.intro}
Classifiers are ubiquitous in today's world. They are used in facial recognition, for military purposes, by autonomous vehicles, and, in this paper, to classify short DNA reads from multiple potential source genomes. Classifier inputs include numerical data, character strings, images and sound clips. The underlying statistical methodologies range from maximum likelihood for mixture models to Bayes methods (as in this paper) to neural networks and deep learning based on labeled ``training data.''

Typically, the input space for a classifier is both high-dimensional and enormous, while the output space is much smaller. When the input space is a graph, insight into the classifier can be gained from understanding its boundary---those points in the input space with one or more neighbors that are classified differently. From the input data perspective, these boundary points can be thought of as fragile, because changing them only ever so slightly changes the output. This usage of the word ``fragile'' differs from that in epidemiology, where fragility is a property of clinical trials datasets that measures how many outcomes must change in order to alter statistical significance of the result. See, e.g., \cite{fragility-wells2021}.  In our experiment, neighbors differ by one nucleotide (a \SNP), which can result from a sequencer error, another data quality problem, or natural polymorphism. Quantifying how much a point differs from its neighbors yields a measure of uncertainty that is computable for any classifier whose input space is a graph.

Understanding the boundary illuminates how the classifier works. Is the boundary large or small? Is its geometry simple or complex? How can it be located and explored? For our case study, the boundary is large, containing as much as 30\% of the input space, not of lower dimension, and complex geometrically. By contrast, boundaries for a ``smooth'' function on a ``continuous'' space---namely, the level sets---are continuous, lower-dimensional manifolds.

The remainder of this paper is organized in the following manner. In Section \ref{sec.experiment}, we describe our experiment: the scientific setting, the data, and the Bayes classifier, as well as introduce two surrogate uncertainty measures. Section \ref{sec.results} contains results for the DNA reads, together with techniques for exploring the boundary. Sections \ref{sec.discussion} and \ref{sec.conclusion} contain discussion and conclusions.

\section{Experimental Setting}\label{sec.experiment}
The problem we address is classifying short DNA sequences generated by \NGSs\ as having arisen from one of three candidate genomes. The dataset we employ is synthetic, so that ground truth---the genome from which each read arises---is known.
The broader scientific context is reference-guided metagenomic assembly---piecing together fragments (reads) of DNA from multiple potential sources into longer sequences called contigs, as performed, for example, by \texttt{MetaCompass} \citep{metacompass2017}, with the assistance of a (reference) database of sequenced candidate genomes. The role of the reference database is to reduce the computational burden of creating the contigs, which is formulated as finding an Eulerian path in a de Bruijn graph \citep{debruijn2023}. The specific problem setting that led to this study involved samples from human digestive and reproductive tracts \citep{amost2024}.

\subsection{Classifier Generalities}\label{subsec.experiment-classifier}
In this paper, a classifier is a function $C$ from a finite input space $\mathcal{I}$ to a finite output space $\mathcal{O}$. Elements of $\mathcal{I}$ are termed inputs, and $C(x) \in \mathcal{O}$ is the decision or result for input $x$. Although it is not a logical necessity, there is little point in constructing or using a classifier unless $|\mathcal{I}| \gg |\mathcal{O}|$, where $|\mathcal{S}|$ is the cardinality of the set $\mathcal{S}$. In our case study, 101 nucleotides in each DNA read with five possible values each yield $|\mathcal{I}| = 5^{101}$, and three candidate genomes yield $|\mathcal{O}| = 3$.

We restrict attention to deterministic classifiers: for an input $x$, there is a unique, reproducible output $C(x)$. In our case study, $C$ is based on maximizing Bayesian posterior probabilities, and there are inherent measures of uncertainty to which we can compare the surrogate measures proposed in Section \ref{subsec.concepts-measures}. Mixture models that represent the effects of (discrete) latent variables are of the same ilk \citep{titterington-mixture85}. By contrast, deep learning and neural network models use a set $\mathcal{T} = \{t_1,\dots,t_m\} \subseteq \mathcal{I}$ of labeled training data to determine tunable parameters of $C$ in such a way that $C(t_j)$ is equal to a known value $d_j$ for each $j$.

We further assume that the input space $\mathcal{I}$ is a loopless, undirected graph, whose edgeset $E$ defines neighboring inputs: $x,y \in \mathcal{I}$ are neighbors if and only $x \neq y$ and $\{x,y\} \in E$. We denote by $\mathcal{N}(x) = \{y: \{x,y\} \in E\}$ the set of neighbors, or neighborhood, of $x$. For ease of interpretation and because it is true in our case study, we further assume that $\mathcal{I}$ is connected: for any $x,y \in \mathcal{I}$, there is a path $(x_1=x, \dots, x_k = y)$ that connects them: $\{x_j, x_{j+1}\} \in E$ for each $j$.

Our output space $\mathcal{O}$ consists of three virus genomes that give rise to (simulated) DNA reads produced by a particular \NGS.

\subsection{Mathematical Preliminaries}\label{subsec.experiment-pre}
To establish notation, a DNA sequence $S$ is a character string chosen from the nucleotide (base) alphabet  $\{\text{A}, \text{C}, \text{G}, \text{T}\}$, where A is adenine, C is cytosine, G is guanine and T is thiamine. At one extreme, $S$ may be an entire genome--for instance, a virus or a chromosome. At another, it may be a read generated by a \NGS. Given a sequence $S$, its length is $|S|$; the $i^{\mathrm{th}}$ base in $S$ is $S(i)$; and the bases from location $i$ to location $j > i$ are $S(i:j)$.

We focus on triplets, whose distributions are 64-dimensional summaries of sequences. That triplets suffice in our case study is discussed below. Additional, more general, discussion appears in \cite{markovstructure-2021}. And, of course, triplets are fundamental biologically because they encode amino acids. Other cases appear in the literature, especially quartets, also referred to as tetranucleotides \citep{pride-tetra-2003, teeling-tetra-classification-2004}. 

The triplet distribution $P_3(\cdot|S)$ of a sequence $S$ is defined as
\begin{equation}
P_3(b_1b_2b_3|S) = \mathrm{Prob}\big\{S(k:k+2) = b_1b_2b_3\big\}
\label{eq.tripletdistribution}
\end{equation}
for each choice of $b_1b_2b_3$, where $b_1$, $b_2$ and $b_3$ are elements of $\{\text{A}, \text{C}, \text{G}, \text{T}\}$, and $k$ is chosen at random from $1, \dots, |S|-2$. An equivalent perspective is that of a second-order Markov chain \citep{markovstructure-2021}. The information contained in $P_3(\cdot|S)$ is the same as that contained in the pair distribution $P_2(\cdot|S)$ and the $16 \times 4$ transition matrix
\begin{equation}
T_3((b_1, b_2), b_3|S) = \mathrm{Prob}\big\{S(k+2) = b_3| S(k) = b_1, S(k+1) = b_2\big\},
\label{eq.transitionmatrix3}
\end{equation}
whose rows are indexed by $(b_1,b_2)$ and columns are indexed by $b_3$, and which gives the distribution of each base conditional on its two predecessors.

\subsection{The Three Genomes and the Reads Dataset}\label{subsec.experiment-genomes-reads}
In our case study, there are three reference genomes: an adenovirus genome of length 34,125, downloaded with the read simulator \texttt{Art}, which we call Adeno; a SARS-CoV-2 genome of length 29,926 contained in a coronavirus dataset downloaded from \NCBI\ in November, 2020, which we call COVID; and a SARS-CoV genome of length 29,751 from the same database, which we call SARS. These are typical lengths for virus genomes. Thus, $\mathcal{O} = \{\text{Adeno}, \text{COVID}, \text{SARS}\}$.


Appendix \ref{app.TD} contains the triplet distributions for these three virus genomes. Measured by Hellinger distance, whose definition is given in (\ref{eq.hellinger}) below, these distributions are very different. The distances are 0.234 for Adeno/COVID, 0.125 for Adeno/SARS and 0.161 for COVID/SARS. To contextualize these values, the .999-quantiles for Hellinger distances of triplet distributions in samples of the same size simulated from each genome are 0.01941755 for adenovirus, 0.02094808 for COVID, and 0.02065539 for SARS. Therefore, none of the triplet distributions is remotely likely to have arisen from another genome, so that classifiers can based on triplet distributions.

The read dataset is the same as in \cite{markovstructure-2021}. We employed the \texttt{Mason\_simulator} software \citep{fu_mi_publications962} to simulate Illumina \NGS\ reads of length 101 from each genome, with approximate 6X coverage. Illumina manufactures \NGSs\ that employ an optical technology; see \texttt{https://www.illumina.com/}. The numbers of reads are 1966, 1996 and 1907, respectively, which sum to 5869. The \texttt{Mason\_simulator} introduces errors in the form of transpositions (\SNPs), insertions, deletions and undetermined bases. The latter are cases in which the sequencer detects that a nucleotide is present, but is unable to determine whether it is A, C, G or T. Following convention, these appear in the simulated reads as ``N,'' and must be accommodated in computations. Some other DNA read datasets contain additional ``partial'' classifications, for instance, a base that is one of two possibilities. Parameters of the \texttt{Mason\_simulator} were set at default values.

\subsection{The Bayes Classifier}\label{subsec.experiment-bayes}
Our classifier $C$ is similar to that in \cite{wang-bayesclass-2007}. For each read $R$, posterior probabilities are calculated for each genome, and the decision for that read is the genome with the maximal posterior probability---the MAP (maximal \emph{a posteriori} probability) estimate.

Specifically, the input space $\mathcal{I}$ is the set of all sequences of length 101 from the set $\{\text{A},\text{C},\text{G},\text{T},\text{N}\}$; the output space is $\mathcal{O} = \{\text{Adeno}, \text{COVID}, \text{SARS}\}$; and $R_1,R_2 \in \mathcal{I}$ are neighbors if and only if $R_1$ and $R_2$ differ by exactly one nucleotide, i.e., their Hamming distance \citep{navarro-string-matching-2001} is equal to one. Therefore, each element of $\mathcal{I}$ has 404 neighbors. Usefully, nature provides a physical interpretation of neighbors as \SNPs.

The three likelihood functions, denoted by $L(\cdot|\text{Adeno})$, $L(\cdot|\text{COVID})$, and $L(\cdot|\text{SARS})$, are calculated from the triplet distributions. To illustrate for adenovirus,
\begin{eqnarray}
\lefteqn{L(R|\text{Adeno}) = P_2(R(1)R(2)|\text{Adeno}) \times T_3((R(1), R(2)), R(3)|\text{Adeno})} \nonumber
\\
& &
\times T_3((R(2), R(3)), R(4)|\text{Adeno}) \times \cdots \nonumber
\\
& &
\times T_3\bigg(\big(R(|R|-2), R(|R|-1)\big), R(|R|)|\text{Adeno}\bigg),
\label{eq.likelihood}
\end{eqnarray}
where $P_2(\cdot|A)$ is the pair distribution defined by analogy with (\ref{eq.tripletdistribution}), and $T_3$ is given by (\ref{eq.transitionmatrix3}). In (\ref{eq.likelihood}), we have ignored Ns for simplicity; when they are present, they lead to sums over all possible bases that could have been the N.

To complete the Bayesian formulation, we assume a uniform prior $\pi_R = (1/3, 1/3, 1/3)$ on $\mathcal{O}$ for each read. We then use Bayes' theorem and the three likelihoods to calculate posterior probability distributions over $\mathcal{O}$, which yield the classifier decisions. For each read $R$, the posterior probability of $x \in \mathcal{O}$ is
\begin{equation}
p(x|R) = \frac{\pi_R(x)L(R|x)}{\pi_R(\text{Adeno})L(R|\text{Adeno}) + \pi_R(\text{COVID})L(R|\text{COVID}) + \pi_R(\text{SARS})L(R|\text{SARS})}.
\label{eq.bayes}
\end{equation}
Finally,
$$
C(R) = \argmax_{x\in \mathcal{O}} p(x|R).
$$
We note that since the $\pi_R$ is uniform, it cancels in (\ref{eq.bayes}), so that $C(R)$ is also the \MLE.

\section{Key Concepts}\label{sec.concepts}
This section introduces the boundary and associated surrogate uncertainty measures.

\subsection{The Boundary}\label{subsec.concepts-boundary}
The boundary $\mathcal{B}$ associated with $C$ is the set of elements of $\mathcal{I}$ that have at least one neighbor classified differently:
\begin{equation}
\mathcal{B} = \{R \in \mathcal{I}: C(R') \neq C(R) \text{ for some } R' \in \mathcal{N}(R)\}.
\label{eq.boundary}
\end{equation}
In principle, the notation should also capture dependence on $C$, but in this case study, $C$ is fixed, so we suppress that dependence.

Whenever two reads are classified differently, any path in $\mathcal{I}$ connecting them must contain at least two (neighboring) boundary points. Below we focus on Hamming paths, which are the shortest. Given reads $R \neq R'$ with Hamming distance $k$ (differing in $k$ of 101 locations), a Hamming path from $R$ to $R'$ simply replaces one differing nucleotide in $R$ at a time by the corresponding nucleotide in $R'$. Therefore, each successive pair on the path are neighbors. Such a path $\big(R_0 = R, \dots, R_k = R'\big)$ has length $k+1$; there are $k!$ of them. The simplest of these paths simply moves left to right in the character strings. Because $C(R) \neq C(R')$, there must be at least one value of $j$ for which $C(R_j) \neq C(R_{j+1})$, in which case, the neighbors $R_j$ and $R_{j+1}$ both belong to $\mathcal{B}$.

To provide initial numerical evidence, Table \ref{tab.confusion} contains the confusion matrix for the Bayes classifier $C$, which is computable because we know the source of each read. The classifier performs well, albeit not spectacularly. Looking only at the columns, Table \ref{tab.confusion} identifies $3,867,933 = 1933 \times 2001$ pairs, the first of which is classified as Adeno and the other as COVID;
$3,740,355 = 1933 \times 1935$ pairs, the first of which is classified as Adeno and the other as SARS;
$3,871,935 = 2001 \times 1935$ pairs, the first of which is classified as COVID and the other as SARS.
Any conclusion about the seemingly large size of these numbers should be tempered by remembering that $|\mathcal{I}| = 5^{101}$.

\begin{table}[ht]
\centering
\caption{Confusion matrix for the Bayes classifier $C$. The correct classification rate is 81.55\%.}
\label{tab.confusion}

\vspace{.1in}
\begin{tabular}{l|rrr|r}
\hline
 & \multicolumn{3}{c|}{Decision} &
\\
Source & Adeno & COVID & SARS & Sum
\\
\hline
  Adeno & 1601 & 115 & 250 & 1966 \\
  COVID & 64 & 1717 & 215 & 1996 \\
  SARS & 268 & 169 & 1470 & 1907 \\
\hline
  Sum & 1933 & 2001 & 1935 & 5869 \\
 \hline
\end{tabular}
\end{table}

\subsection{Surrogate Uncertainty Measures}\label{subsec.concepts-measures}
The Bayes classifier $C$ possesses two inherent and related measures of uncertainty, the maximum posterior likelihood and the entropy of the posterior distribution. 
Other classifiers may lack such measures, although there might be informal measures of ``confidence.'' For instance, partition models often characterize confidence in terms of the homogeneity of terminal nodes, which is a useful but not fully principled measure, since it does not result from a statistical model.


We now introduce three surrogate uncertainty measures that are computable for any classifier operating on the same input space $\mathcal{I}$. Two of them, for a read $R$, are functions of the distribution on $\mathcal{O}$ of $\{C(R'): R' \in \mathcal{N}(R)\}$.

The first measure, Boundary Status, of a read $R$ is the number of genomes other than that assigned to $R$ itself appearing among the outputs for its 404 neighbors. Thus, $\text{BS}(R) = 0$ means that all neighbors are classified the same as the read, i.e., $R$ is not on the boundary; $\text{BS}(R) = 1$ means that one of the other genomes appears among the neighbors; and $\text{BS}(R) = 2$ means that both other genomes do.

The second measure, Neighbor Similarity, quantifies the extent to which the neighbors of a read have the same decision it does, using Hellinger distance \citep{nikulin-hd-2010}. The Hellinger distance between distributions $p$ and $q$ on a finite set $S$ is given by
\begin{equation}
H(p, q) = \sqrt{\frac{1}{2}\sum_{x \in S}\bigg(\sqrt{p(x)} - \sqrt{q(x)}\bigg)^2}.
\label{eq.hellinger}
\end{equation}
Hellinger distance is a metric, and takes values between zero (if and only if $p = q$) and one (when $p$ and $q$ have disjoint supports).

Neighbor Similarity is then defined for a read $R$ as
\begin{equation}
\text{NS}(R) = 1 - H(q_R, q_{N(R)}),
\label{eq.ns}
\end{equation}
where $H$ denotes Hellinger distance, $q_R$ is the degenerate probability distribution on $\mathcal{O}$ concentrated on $C(R)$, and $q_{N(R)}$ is the probability distribution on $\mathcal{O}$ of $\{C(R'): R' \in \mathcal{N}(R)\}$. Therefore,
$$
\mathcal{B} = \{R \in \mathcal{I}: \text{NS}(R) < 1\}.
$$

Unlike $\text{BS}$, $\text{NS}$ takes account only of how many neighbors differ, not their values.
Our principal proposed interpretation of $\text{NS}(R)$ is as uncertainty regarding the classification of $R$. If $\text{NS}(R) = 1$, all neighbors of $R$ have the same decision as $R$, and, intuitively, we would be more certain of the decision for $R$. At the other extreme, if $\text{NS}(R)$ were zero, changing any single nucleotide of $R$ changes the decision, in which case we would be highly uncertain about $C(R)$. 

However, as discussed further in Section \ref{subsubsec.other-datasets}, we must distinguish certainty from correctness. In some regions of $\mathcal{I}$, the Neighbor Similarity may be high because ``everything'' is classified the same way, correctly or not.

Both $\text{BS}$ and $\text{NS}$ can be calculated for any classifier whose input space has the graph structure described in Section \ref{sec.experiment}, and their values returned along with the classifier decision. The burden is computational: the classifier must be run on all neighbors of the input. See Section \ref{sec.discussion} for further discussion.

A third surrogate measure is more speculative, because we lack a demonstrably efficient method for calculating it. It is the Distance from the Boundary: $\text{DB}(R)$ is the minimum value of $k$ for which there exists a path $(R_0 = R, \dots, R_k)$ in $\mathcal{I}$ such that $R_k \in \mathcal{B}$. Then, the interpretation is that the larger $\text{DB}(R)$, the greater the certainty regarding $C(R)$, because the more $R$ would have to be ``mutated'' in order to change $C(R)$. It is easy to obtain upper bounds on $C(R)$: for any other read $R'$ with $C(R') \neq C(R)$, $\text{DB}(R)$ is at most the Hamming distance between $R$ and $R'$. The unresolved problem is exact computation of $\text{DB}(R)$. Also, the ubiquity of the boundary suggests that $\text{DB}$ may lack discriminatory power not already present in $\text{BS}$ and $\text{NS}$.

\section{Results}\label{sec.results}
All analyses reported here were performed using the \textsf{R} Language and Environment for Statistical Computing \citep{R}.

\subsection{DNA Reads}\label{subsec.results-reads}
Table \ref{tab.boundary-crosstabs} contains cross-tabulations of Boundary Status for the 5869 reads as a function of (1) read source, (2) the classifier decision and (3) correctness of the classifier decision. We observe that 1791 of the 5869 reads, or 30.52\%, lie on the boundary $\mathcal{B}$, which is the first concrete evidence that $\mathcal{B}$ is not ``thin.'' Of these 1791 boundary reads, 190 (3.24\%) have Boundary Status equal to 2.

Looking in more detail, Figure \ref{fig.reads-neighbordist} shows the neighbor distributions over $\mathcal{O}$ for all the reads. In it, three-dimensional probabilities (barycentric coordinates) are converted to Cartesian coordinates, as points in an equilateral triangle. Pure Adeno, in the sense that $\text{Prob}(\text{Adeno}) = 1$, is the top vertex, pure COVID is the lower left vertex, and pure SARS is the lower right vertex. Because we know the sources of the reads, we create separate displays for each source. The white diamond in each scatterplot is the centroid of the probabilities it contains, and all three centroids lie close to the source. A read on an edge has neighbors of two kinds, one of which is that of the read itself. In theory, a read could have $\text{BS}(R) = 2$  without neighbors of its same type, but this case of a read classified differently from {all of its neighbors does not occur in our dataset. The minimum number of concordant neighbors is 11. A read in the interior of the triangle has neighbors with all three decisions.

The topmost table in Table \ref{tab.boundary-crosstabs} shows Boundary Status as a function of read source, so that its row sums match those in Table \ref{tab.confusion}.  The distribution of Boundary Status is not uniform across the three sources: boundary percentages are 30.0\% for Adeno reads, 22.1\% for COVID reads and 40.0\% for SARS reads. The $p$-value for the $\chi^2$ test on this table is less than $10^{-16}$, so the differences are significant. SARS reads are more likely than others to lie on the boundary.

The middle table in Table \ref{tab.boundary-crosstabs} provides a complementary view of Boundary Status as a function of classifier decision, so that its row sums match the column sums in Table \ref{tab.confusion}. Here also, the $\chi^2$ test is massively significant, with $p$-value less than $10^{-16}$. SARS remains the outlier: relatively more reads classified as SARS lie on the boundary than for Adeno or COVID. We do not have a biological hypothesis for this.

The bottom table in Table \ref{tab.boundary-crosstabs} shows Boundary Status as a function of correctness of the classifier decision, and begins to get at the heart of the matter. Of incorrectly decided reads, 64.66\% lie on the boundary, while only 22.81\% of correctly decided reads lie on the boundary! The classifier is struggling in the vicinity of $\mathcal{B}$, and one rightly should be less confident of such decisions. The $\chi^2$ statistic for this table is 756.86, more than five times that for source (topmost table), and three times that for decision (middle table). Moreover, the Kolmogorov-Smirnov test for the two \ECDFs\ is massively significant, with a p-value less than $10^{-16}$: Boundary Status is higher for incorrect decisions than for correct ones.

\begin{figure}[ht]
\centering
\includegraphics[width = 5in]{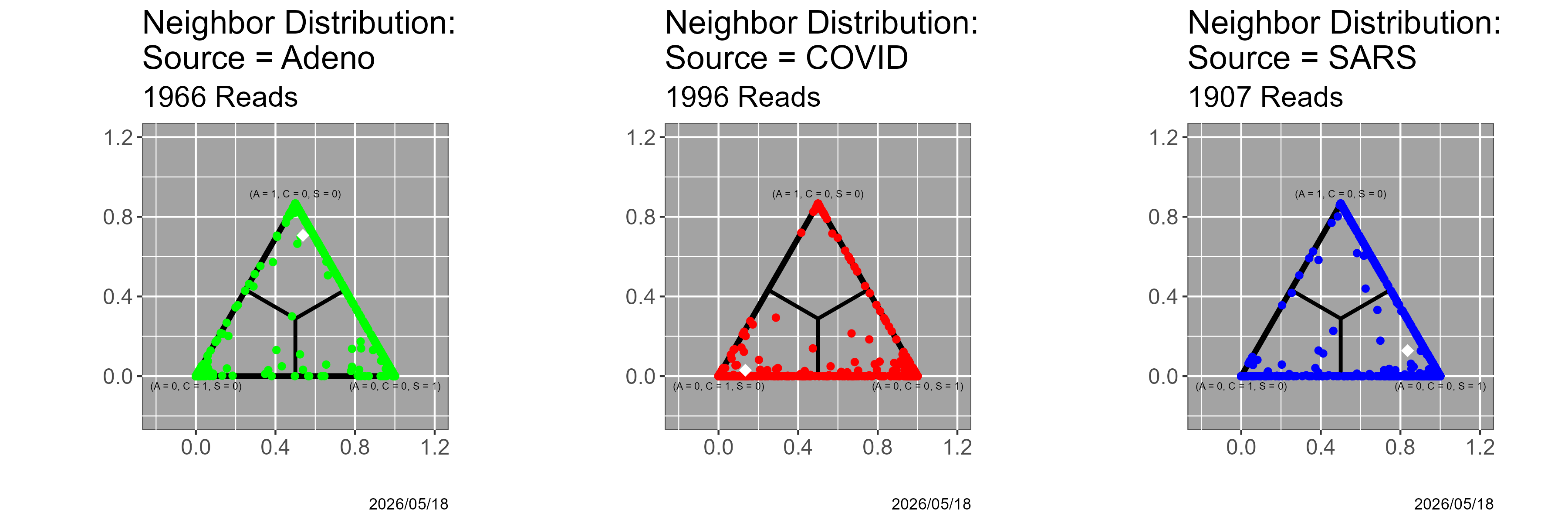}
\caption{Neighbor distributions for the 5869 reads, by source. The geometry is explained in the text.}
\label{fig.reads-neighbordist}
\end{figure}

\begin{table}[ht]
\centering
\caption{\emph{Top:} Cross-tabulation of read source and Boundary Status. \emph{Middle:} Cross-tabulation of read decision and Boundary Status. \emph{Bottom:} Cross-tabulation of decision correctness and Boundary Status.}
\label{tab.boundary-crosstabs}

\vspace{.1in}
\begin{tabular}{r|rrr|r}
  \hline
  & \multicolumn{3}{c|}{Boundary Status} & \\
  Source & 0 & 1 & 2 & Sum \\
  \hline
  Adeno & 1378 & 526 & 62 & 1966 \\
  COVID & 1554 & 375 & 67 & 1996 \\
  SARS & 1146 & 700 & 61 & 1907 \\
  \hline
  Sum & 4078 & 1601 & 190 & 5869 \\
   \hline
\end{tabular}

\vspace{.25in}\begin{tabular}{r|rrr|r}
  \hline
  & \multicolumn{3}{c|}{Boundary Status} & \\
  Decision & 0 & 1 & 2 & Sum \\
  \hline
  Adeno & 1408 & 491 & 34 & 1933 \\
  COVID & 1575 & 345 & 81 & 2001 \\
  SARS & 1095 & 765 & 75 & 1935 \\
  \hline
  Sum & 4078 & 1601 & 190 & 5869 \\
  \hline
\end{tabular}

\vspace{.25in}\begin{tabular}{r|rrr|r}
  \hline
  & \multicolumn{3}{c|}{Boundary Status} & \\
  Correct? & 0 & 1 & 2 & Sum \\
  \hline
  No & 382 & 598 & 101 & 1081 \\
  Yes & 3696 & 1003 & 89 & 4788 \\
  \hline
  Sum & 4078 & 1601 & 190 & 5869 \\
   \hline
\end{tabular}
\end{table}

In parallel to Table \ref{tab.boundary-crosstabs}, Figure \ref{fig.ns-ecdfs} shows the distribution of Neighbor Similarity as a function of read source, read decision and decision correctness. In particular, these \ECDFs\ make clear how many reads on the boundary, for which $\text{NS}(R) < 1$. The qualitative messages mirror those in Table \ref{tab.boundary-crosstabs}.

\begin{figure}[ht]
\centering
\includegraphics[width=2in]{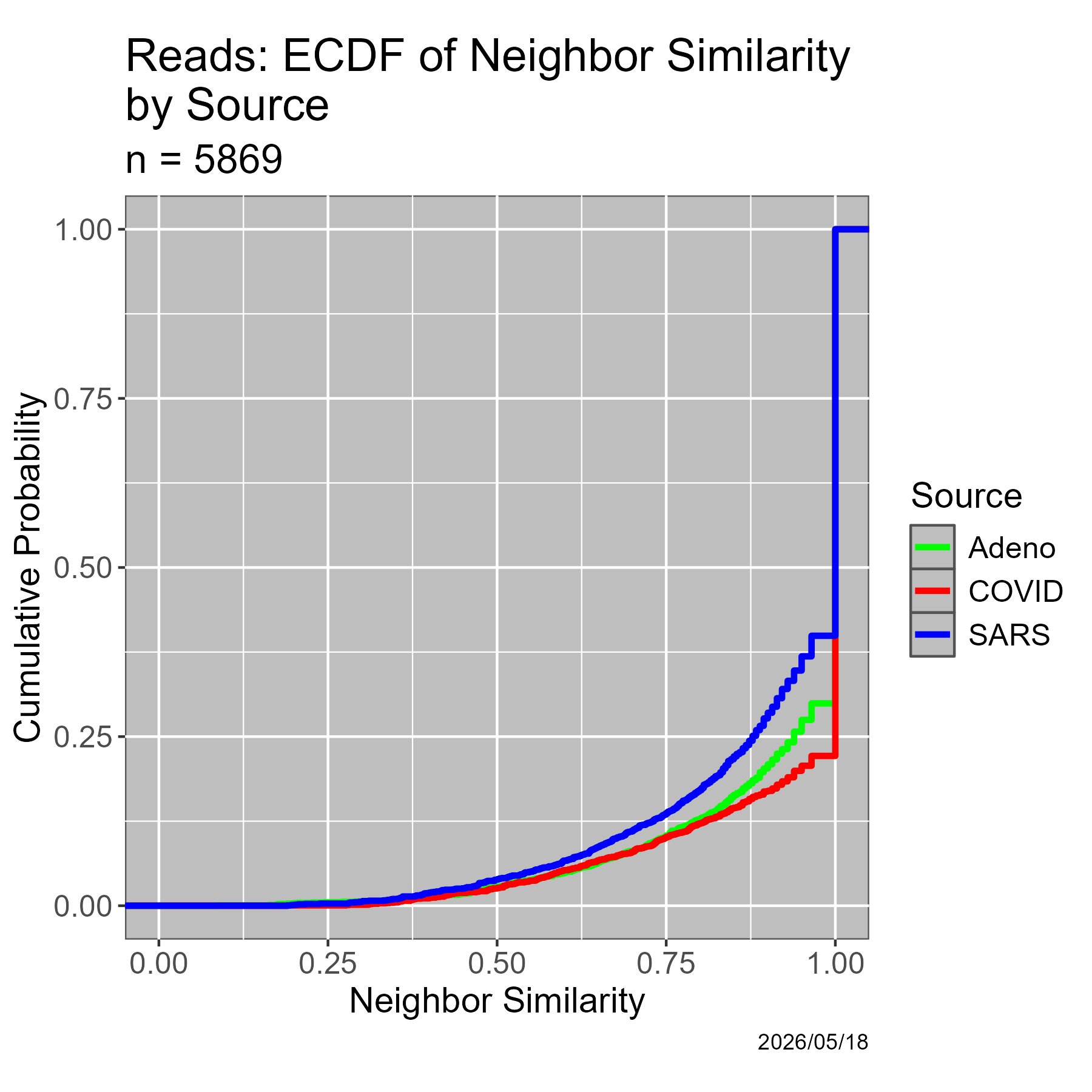}\hspace{.1in}\includegraphics[width=2in]{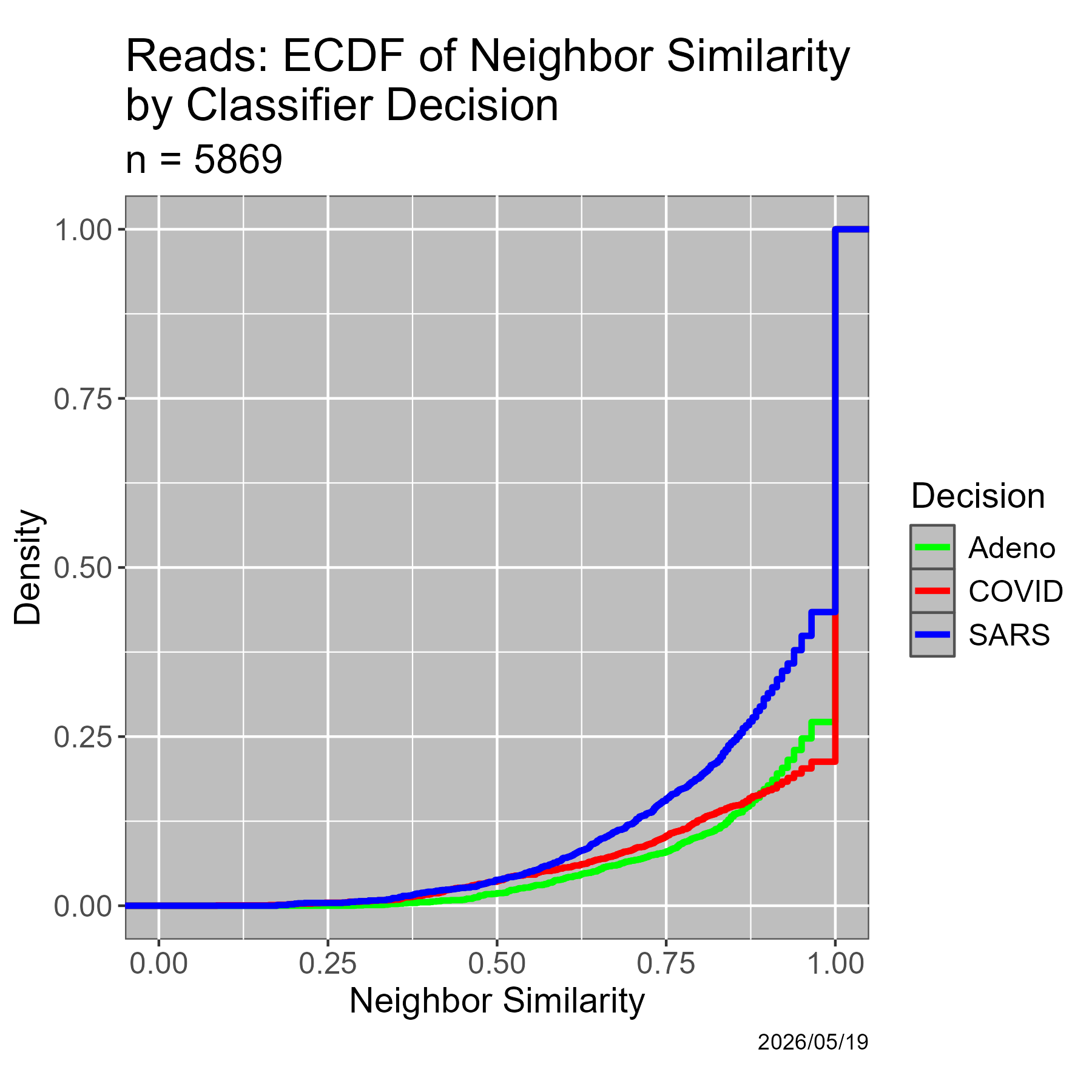}

\vspace{.2in}\includegraphics[width=2in]{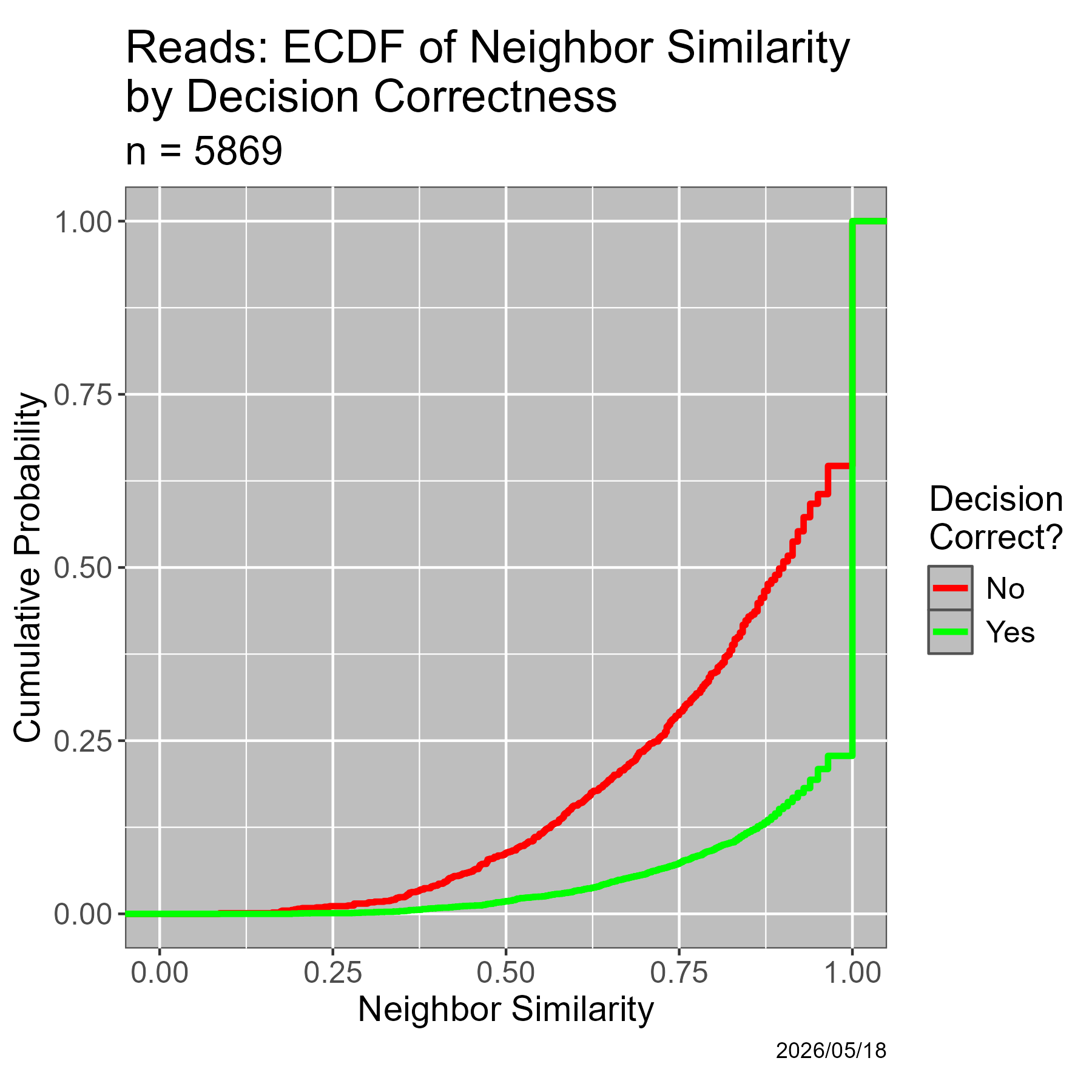}
\caption{ECDFs of Neighbor Similarity. \emph{Upper left:} by read source. \emph{Upper right:} by classifier decision. \emph{Bottom:} by decision correctness. In each panel, the $x$-axis is Neighbor Similarity and the $y$-axis is cumulative probability.}
\label{fig.ns-ecdfs}
\end{figure}

Neighbor Similarity is more burdensome computationally than simply running the classifier. However, the computation is infinitely parallelizable: classifier evaluation for one neighbor is completely independent of that for other neighbors. An alternative strategy is to sample neighbors, thereby generating estimates for Neighbor Similarity. Figure \ref{fig.sampling} provides some evidence that sampling can be effective. It is based on the entire 5869-element read dataset. The $x$-axis is the number of reads sampled; the order was randomized for each read. The $y$-axis is more complicated. For each read $R$ and sample size $k$ we calculated a partial Neighbor Similarity $\text{NS}(R, k)$ using (\ref{eq.ns}). Then for each $k$, we fitted a linear model with $\text{NS}(\cdot, 404) = \text{NS}$ as response and $\text{NS}(\cdot, 1), \dots,  \text{NS}(\cdot, k)$ as predictors. Finally, the $y$-axis contains values of relative root mean squared error (RRMSE---the square root of the mean squared error for the linear model divided by the mean response) for each model. From the figure, an RRMSE of 5\% requires only 20 sampled neighbors, and there is minimal benefit from sample sizes exceeding 80. How general these results are, and in particular how values scale with respect to problem size, is not clear.

\begin{figure}[ht]
\centering
\includegraphics[width=2in]{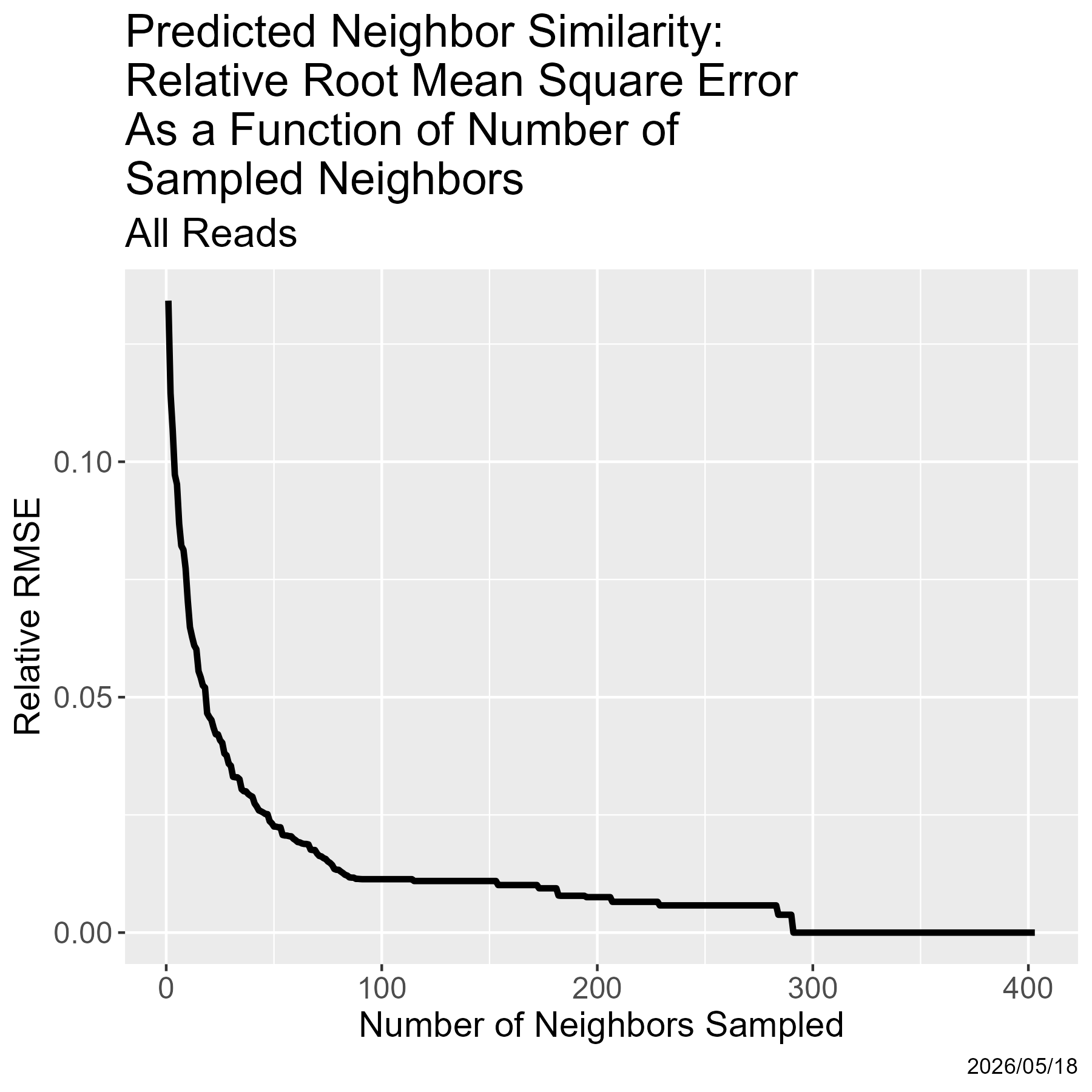}
\caption{Relative root mean squared error (RRMSE) in estimating $\text{NS}$ from samples of neighbors. The $x$-axis is the number of samples and the $y$-axis is RRMSE.}
\label{fig.sampling}
\end{figure}

\subsection{Relationship of Neighbor Similarity to Inherent Measures of Uncertainty}\label{subsec.inherent}
We next investigate the relationships between Neighbor Similarity and the two inherent measures for $C$: $\text{MP}(\cdot) $---the maximum of the three posterior probabilities, and $\text{PE}(\cdot)$---the entropy of the posterior distribution.

Always, $\text{MP}(R) \geq 1/3$, and the closer $\text{MP}(R)$ is to 1, the more certain $C$ is of the decision for $R$. Figure \ref{fig.maxpost-vs-ns} depicts the relationship between $\text{MP}$ and $\text{NS}$ as a function of classifier decision. Because one ultimate goal is to employ $\text{NS}$ in contexts where neither $\text{MP}$ nor a model-derived analog is available, it would not have been appropriate to use read source there, rather than classifier decision. The correlation between $\text{MP}$ and $\text{NS}$ is $0.8488$; the Spearman correlation is 0.7945. Disaggregated correlations are 0.8784 for Adeno, 0.8502 for COVID and 0.8400 for SARS, and corresponding Spearman values are 0.8784, 0.8502, and 0.8310.

We can construct good statistical models to predict $\text{MP}(R)$ from $\text{NS}(R)$ and $C(R)$, of which we present two. The first is a fully saturated quadratic model reflecting the curvature evident in Figure \ref{fig.maxpost-vs-ns}:
\begin{equation}
\text{MP} = \alpha_{C(R)}\text{NS}(R)^2 + \beta_{C(R)}\text{NS}(R) + \gamma_{C(R)}.
\label{eq.regmodel}
\end{equation}
As the notation suggests, the process is equivalent to fitting three separate models to datasets constructed by filtering on $C(R)$. The adjusted coefficient of determination for the model is 0.8341, indicating good fit. Figure \ref{fig.maxpost-pred-act-reg} shows actual and predicted values of $\text{MP}$ by classifier decision. Much of the imperfect fit results from inability to predict the continuous variable $\text{MP}$ within sets of reads having the same value of $\text{NS}$ (especially those with $\text{NS} = 1$).

\begin{figure}[ht]
\centering
\includegraphics[width=5in]{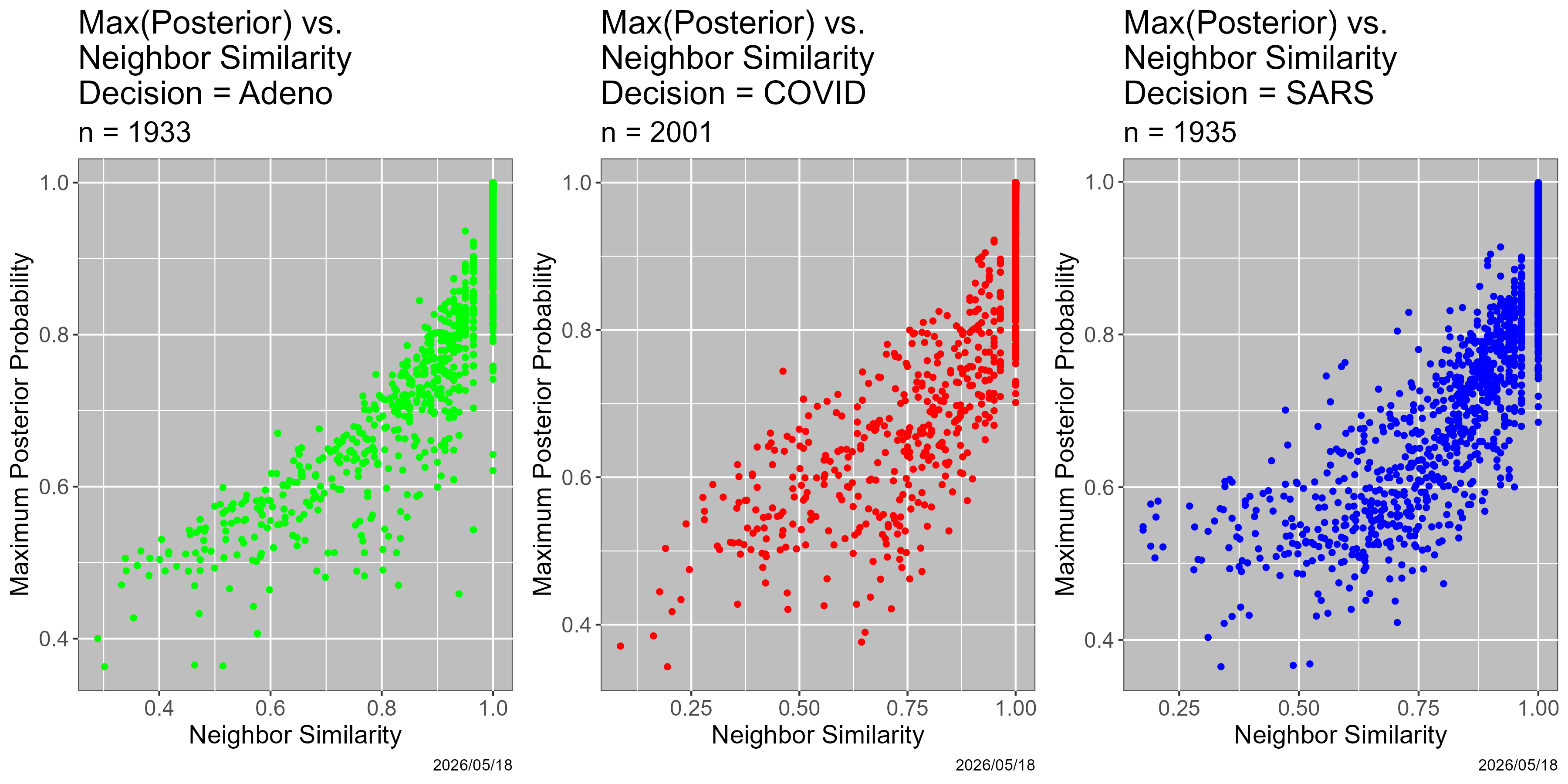}
\caption{Scatterplots of $\text{MP}$ versus $\text{NS}$, by classifier decision. \emph{Left:} decision = Adeno, \emph{Center:} decision = COVID. \emph{Right:} decision = SARS. In each panel, the $x$-axis is Neighbor Similarity and the $y$-axis is MP.}
\label{fig.maxpost-vs-ns}
\end{figure}

\begin{figure}[ht]
\centering
\includegraphics[width=5in]{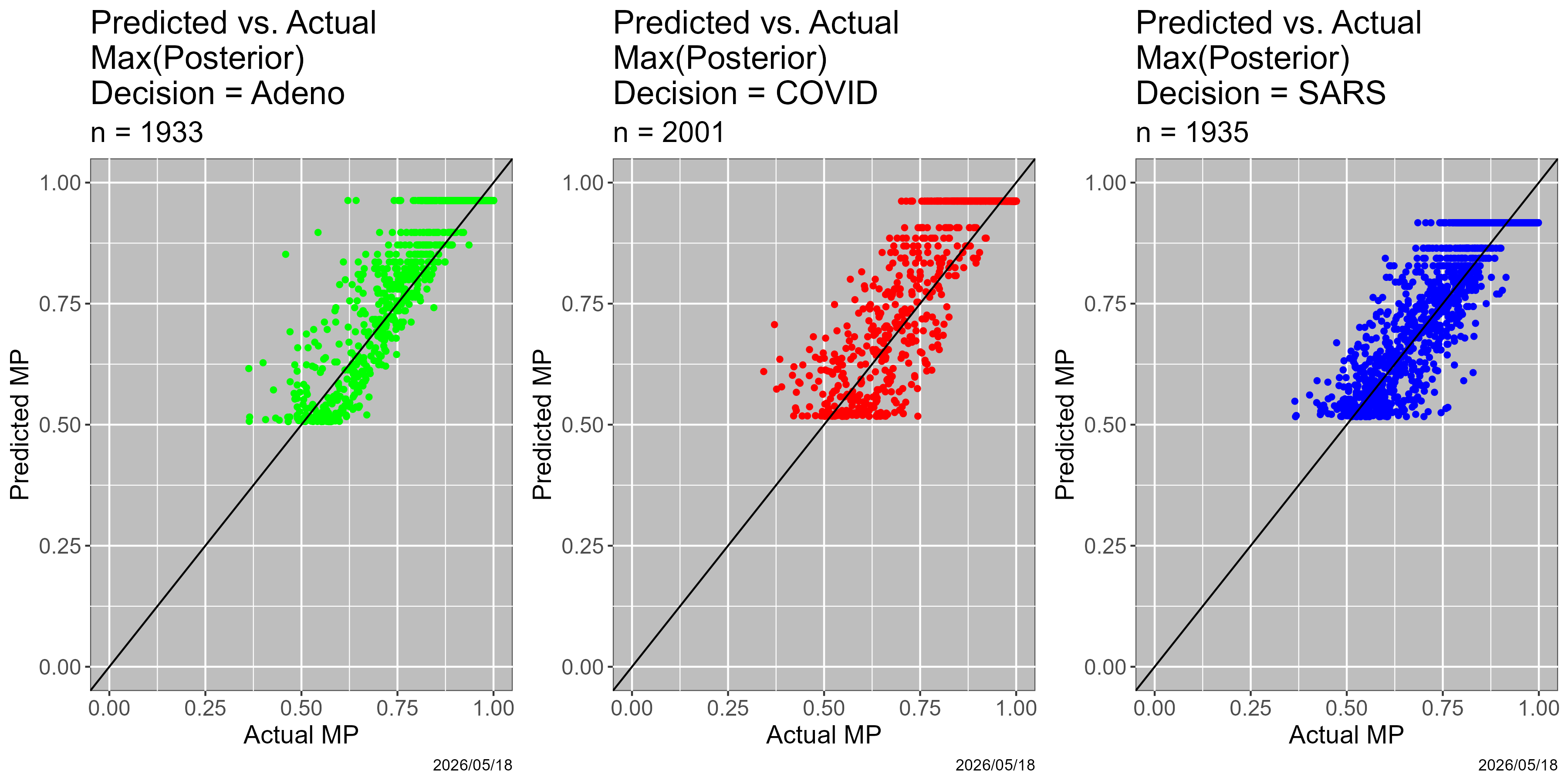}
\caption{For the quadratic regression model, scatterplots of predicted $\text{MP}$ versus actual $\text{MP}$, by classifier decision. \emph{Left:} decision = Adeno, \emph{Center:} decision = COVID. \emph{Right:} decision = SARS.}
\label{fig.maxpost-pred-act-reg}
\end{figure}

Figure \ref{fig.tree-MP} shows a partition model---in this case, a regression tree \citep{cart17, friedmanhastietibshirani-2001}. The tree has been pruned to only seven terminal nodes using standard heuristics that trade off predictive accuracy for model complexity. The mean squared error is slightly superior to that for the regression model---0.003313057 as compared to 0.003794415.

\begin{figure}[ht]
\centering
\includegraphics[width=5in]{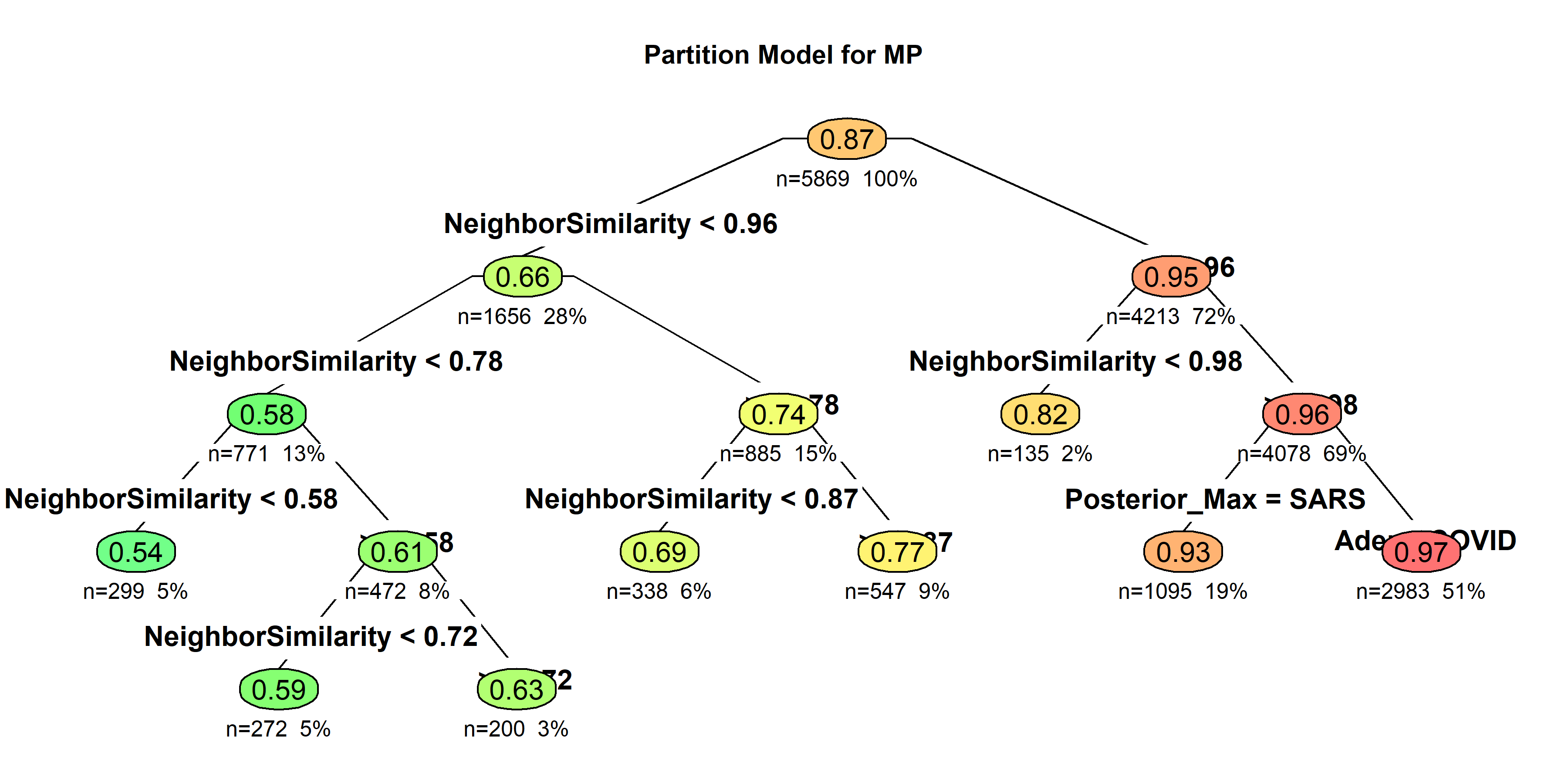}
\caption{Partition model for $\text{MP}(R)$ using $\text{NS}(R)$ and $C(R)$ as predictors. Values in each node are means of $\text{MP}$; numbers below nodes are counts and percentages; edges are labeled by the predictor on which the split occurs and the associated value.}
\label{fig.tree-MP}
\end{figure}

The second inherent measure of uncertainty for $C$ is posterior entropy $\text{PE}$.  The correlations between $\text{NS}$ and $\text{PE}$ are -0.7167 (aggregated), -0.7437 (Adeno), -0.7280 (COVID) , and -0.6964 (SARS). Analogously to Figure \ref{fig.maxpost-vs-ns}, Figure \ref{fig.pe-vs-ns} contains scatterplots of $\text{PE}$ versus $\text{NS}$ by classifier decision. Clearly, $\text{PE}$ is predicted less well by $\text{NS}$ and classifier decision than is $\text{MP}$. For a quadratic regression model analogous to that in (\ref{eq.regmodel}), the adjusted $R^2$ is 0.6928. The corresponding partition model, shown in Figure \ref{fig.tree-PE}, identifies which variables and interactions are most relevant. Again, the mean squared error is slightly lower than for the quadratic regression model.

\begin{figure}[ht]
\centering
\includegraphics[width=5in]{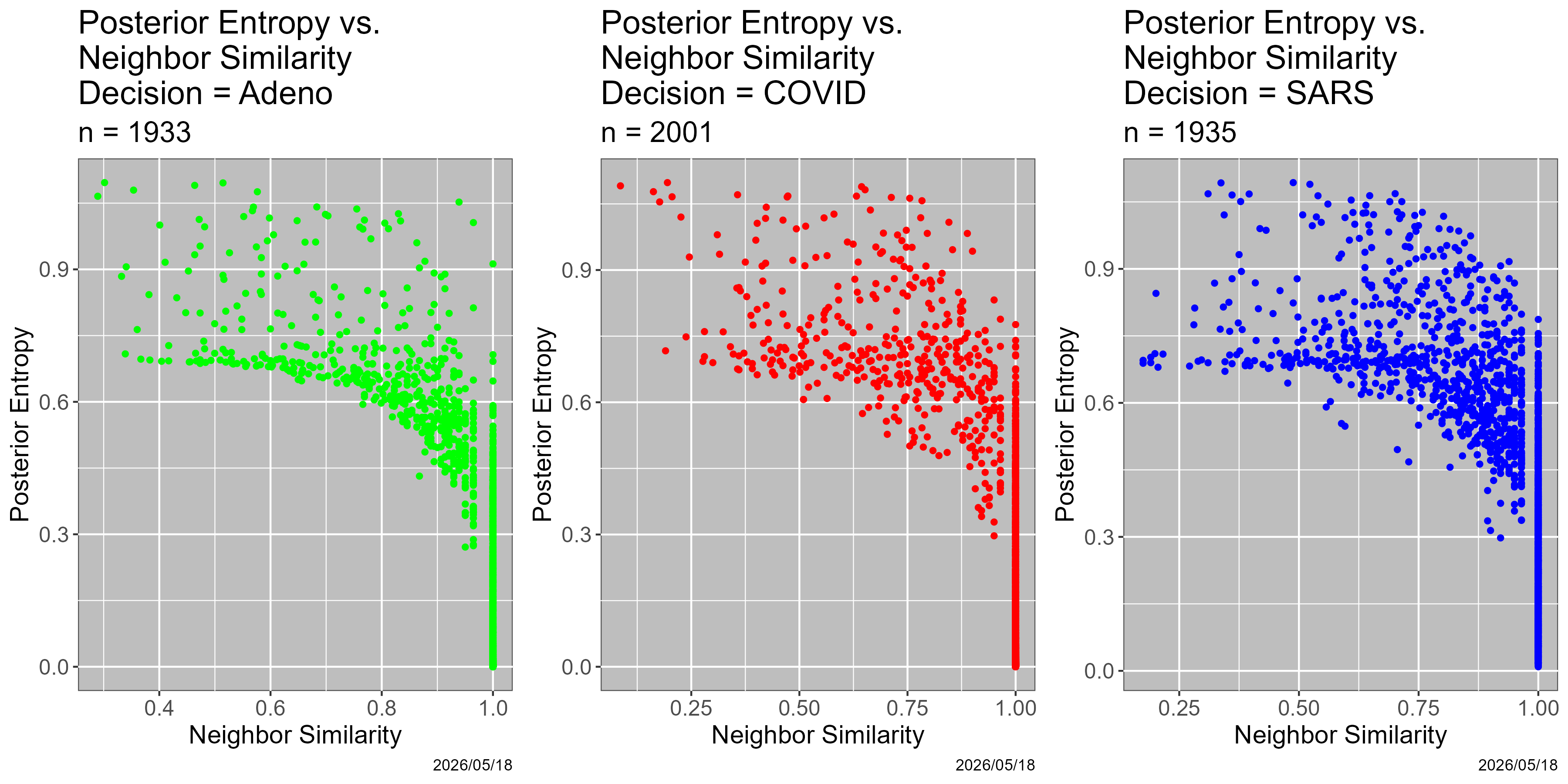}
\caption{Scatterplots of $\text{PE}$ versus $\text{NS}$, by classifier decision. \emph{Left:} decision = Adeno, \emph{Center:} decision = COVID. \emph{Right:} decision = SARS.}
\label{fig.pe-vs-ns}
\end{figure}

\begin{figure}[ht]
\centering
\includegraphics[width=5in]{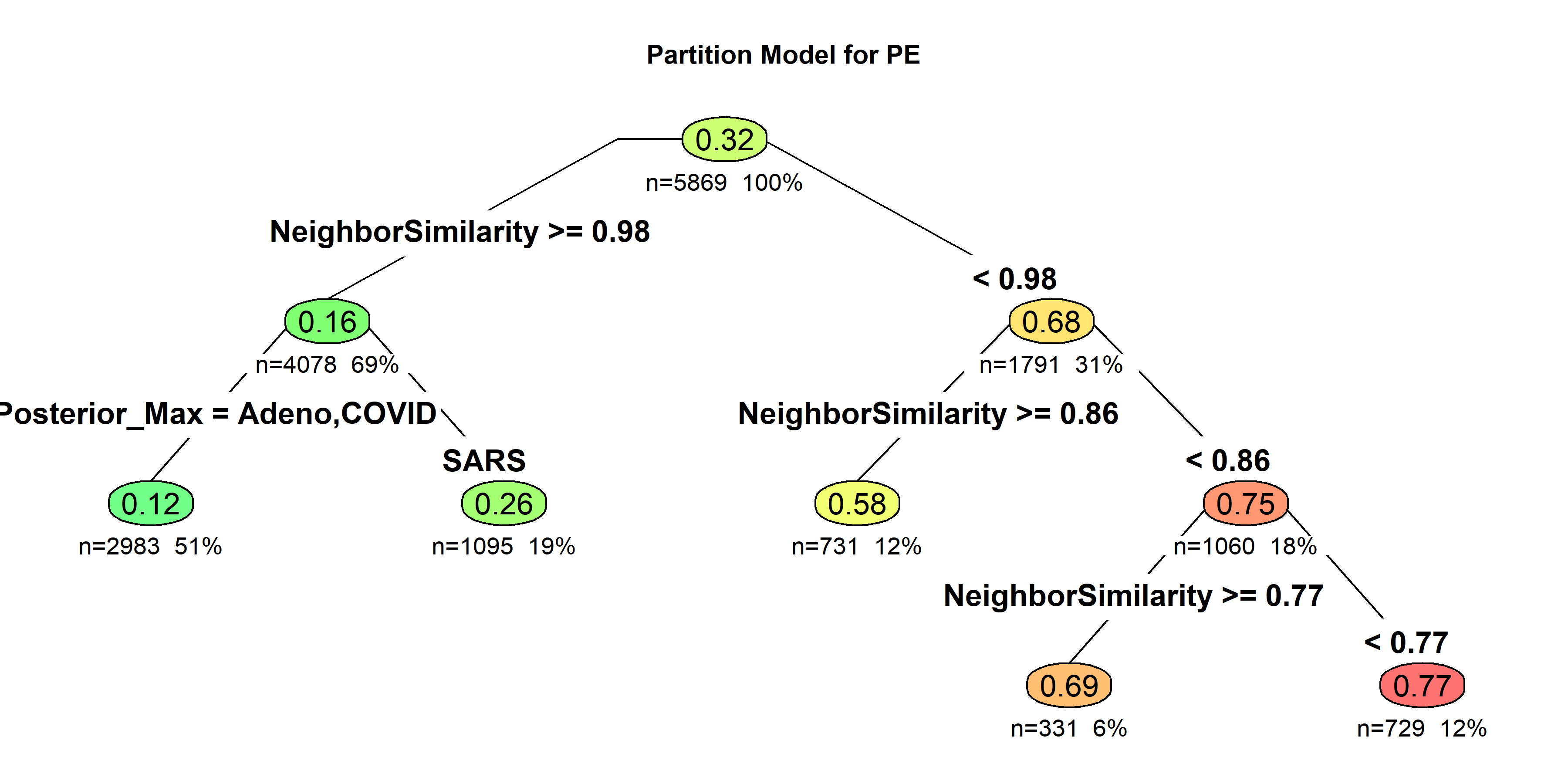}
\caption{Partition model for $\text{PE}(R)$ using $\text{NS}(R)$ and $C(R)$ as predictors. Values in each node are means of $\text{PE}$; numbers below nodes are counts and percentages; edges are labeled by the predictor on which the split occurs and the associated value.}
\label{fig.tree-PE}
\end{figure}

\subsection{Exploring the Boundary}\label{subsec.results-explore}
We present here three strategies for locating the boundary $\mathcal{B}$ and exploring its structure.

The first and simplest method for locating $\mathcal{B}$ has been mentioned in Section \ref{subsec.concepts-boundary}: if $C(R) \neq C(R')$, then any path from $R$ to $R'$ contains at least two (neighboring) elements of $\mathcal{B}$. Within the reads dataset, there are 11,480,223 such pairs. If we eliminate duplicates and for each pair consider only the Hamming path that in left-to-right order replaces each nucleotide in $R$ differing from the corresponding nucleotide in $R'$ by the latter, then we have a mechanism for locating boundary pairs. Based on a random sample of 25 origin reads, but considering all other reads with different decisions, 1,242,162 boundary points were identified. There is significant variability, however, as shown in Table \ref{tab.minimalHamming}. Moreover, the table shows that paths originating from reads classified as COVID differ from those originating at reads classified as Adeno or SARS.

\begin{table}[ht]
\centering
\caption{Results for Hamming paths from 25 randomly sampled reads to all other reads with differing classifier decisions.}
\label{tab.minimalHamming}

\vspace{.1in}
\begin{tabular}{r|lrr}
  \hline
  SampleID & Decision & Boundary Points & Classifier Evaluations \\
  \hline
  958 & Adeno & 71646  & 296286  \\
  1283 & SARS & 59530  & 298994  \\
  256 & Adeno & 50388  & 294923  \\
  1137 & SARS & 32052  & 303790  \\
  638 & SARS & 75160  & 296341  \\
  4905 & Adeno & 57804  & 302409  \\
  2914 & Adeno & 42374  & 292198  \\
  513 & COVID & 63228  & 289336  \\
  2160 & COVID & 52038  & 292711  \\
  2835 & COVID & 58806  & 293286  \\
  1386 & Adeno & 31784  & 304305  \\
  3584 & Adeno & 40992  & 296839  \\
  1912 & SARS & 34274  & 296172  \\
  4857 & Adeno & 31002  & 302553  \\
  3713 & Adeno & 37630  & 292638  \\
  4196 & Adeno & 32332  & 300289  \\
  5113 & Adeno & 68118  & 293587  \\
  4090 & SARS & 43252  & 297002  \\
  1981 & Adeno & 49500  & 293600  \\
  2666 & Adeno & 50380  & 295404  \\
  2153 & COVID & 60832  & 295370  \\
  2674 & SARS & 52446  & 295611  \\
  351 & Adeno & 75608  & 301328  \\
  5154 & SARS & 37546  & 297293  \\
  1940 & Adeno & 33440  & 303009  \\
   \hline
\end{tabular}
\end{table}

Of course, whenever $C(R) \neq C(R')$, there are multiple Hamming paths connecting $R$ and $R'$. Initial experiments have shown that a strategy of using multiple Hamming paths is not materially more effective than using one Hamming path in terms of unique boundary pairs identified relative to computational effort. The reason, we believe, is that multiple paths from $R$ to $R'$ often cross $\mathcal{B}$ at the same places.

A second strategy, motivated by Markov chain Monte Carlo (MCMC) methods used to investigate large spaces of contingency tables \citep{diaconissturmfels98}, is to perform random walks originating from sequences $x \in \mathcal{I}$, moving to a neighboring sequence at each step, and checking whether $\mathcal{B}$ has been crossed, i.e., whether the classifier decision has changed. Were $\mathcal{B}$ ``thin,'' this strategy of wandering randomly would be horribly ineffective. But, because $\mathcal{B}$ is not thin, it works reasonably well. To illustrate, we chose a random sample of 100 reads for each decision, and ran random walks of length 2000 steps originating at each. Keeping in mind that each boundary pair comprises two boundary points, 35,352 boundary points were found, of 600,300 sequences visited.

The hypothesis that this process would be more efficient if started from a boundary point is not confirmed. We repeated the experiment starting from 300 sequences known to be boundary points, again spread uniformly over the three decisions. The improvement over random walks starting from random sequences is minimal; indeed, fewer boundary points---35,238---were identified. We interpret this as further confirmation that the boundary is not thin, so that starting from it yields little benefit. Note that none of the 600 random walks failed to cross the boundary, so it is never far away. Summary statistics appear in Table \ref{tab.rwSummary}.

\begin{table}[ht]
\centering
\caption{Comparison of summary statistics for 300 random walks originating at reads and for 300 random walks originating at boundary points.}
\label{tab.rwSummary}

\vspace{.1in}
\begin{tabular}{l|rrr}
\hline
Origins & Minimum & Mean & Maximum
\\
\hline
Reads & 28 & 117.77 & 274
\\
Boundary Points & 16 & 117.46 & 280
\\
\hline
\end{tabular}
\end{table}

From Table \ref{tab.exploreboundary}, both versions of random walks are less efficient than Hamming paths, as measured by the number of boundary points identified divided by the number of classifier evaluations, which is the primary computational expense of boundary exploration. But, these results again confirm the omnipresence of $\mathcal{B}$.

Note that in our implementation both Hamming paths and random walks may pass through boundary points without identifying them, which can happen because (for computational efficiency) each point in the path is compared only to its predecessor rather than all of its neighbors.

Finally, we present random walk-like methods for ``crawling the boundary'' once it has been reached.  We simulated 100 random walks starting at boundary points, but constrained to move only to adjacent and not previously visited boundary points. Each crawl was allowed to continue for 250 steps, or until it was not possible to move to another and not previously visited boundary point. The fact that such points exist further illuminates the geometric complexity of the boundary: it is ''hairy'' in the sense that the points from which no continuation is possible are tips of hairs.

While insightful, crawling the boundary is not efficient. Of the 100 random walks, 64 did not terminate within 250 steps, indicating that the boundary $\mathcal{B}$ is significantly contiguous. See also Section \ref{subsec.connected}. The lengths of crawls that did terminate range from 5 to 242, with no discernible structure to their distribution. The 36 crawls that did terminate are the hairs mentioned in the preceding paragraph---points at which continuing on the boundary is possible only by backtracking. While in the minority, they are nevertheless not rare.

Beyond this, the boundary is convoluted: all three decisions are present in 58 of the 100 crawls. Put differently, if one were to think of three boundaries---an Adeno--COVID boundary, an Adeno--SARS boundary, and a COVID--SARS boundary---they cross each other frequently, as evidenced by the 190 reads with neighbors of all three types. In all other crawls, the decisions must and do alternate, indicating that some segments of the boundary lie between two decisions.

The inefficiency of crawling the boundary, which is apparent in Table \ref{tab.exploreboundary}, is partly an artifact of our computational implementation. At each step, we classify all neighbors before determining if it is possible to proceed. For related discussion, see Section \ref{subsec.connected}.



\begin{table}[ht]
\centering
\caption{Summary of methods for exploring the boundary.}
\label{tab.exploreboundary}

\vspace{.1in}
\begin{tabular}{l|rrr}
\hline
Method & Classifier Evaluations & Boundary Points & Efficiency
\\
\hline
Hamming Paths & 7,425,274 & 1,242,162 & 16.73\%
\\
Random Walks, Random Origin & 600,300 & 34,932 & 5.82\%
\\
Random Walks, Boundary Origin & 600,300 & 25,238 & 5.87\%
\\
Crawling Boundary & 8,146,256 & 20,164 & 0.25\%
\\
\hline
\end{tabular}
\end{table}

\subsection{Do Boundary Points Differ from Other Reads?}\label{subsec.results-BPdiffer}
Here we address a natural question: do the 1791 boundary reads in our dataset differ from the 4078 other reads? Figure \ref{fig.mds} provides initial insight. It contains scatterplots of the four components of a \MDS\, based on \cite{kruskal-mds-1978}, of the triplet distributions, colored by Boundary Status. The most striking feature is the concentration of values of Component 1 for reads $R$ with $\text{BS}(R) = 2$, which is reinforced by the upper left-hand panel in Figure \ref{fig.mds-ecdf}. This latter figure contains plots of the ECDFs of each of the four components, disaggregated by Boundary Status; the colors match those in Figure \ref{fig.mds}. In fact because of overplotting in Figure \ref{fig.mds}, there is additional structure that is less visible in it than in Figure \ref{fig.mds-ecdf}. Notably, there are many instances where one ECDF lies entirely below another, including Component 1 ($\text{BS} = 2$) below Component 1 ($\text{BS} = 1$); effectively, Component 3 ($\text{BS} = 0$) below Component 3 ($\text{BS} = 1$) below Component 3 ($\text{BS} = 2$); Component 4 ($\text{BS} = 0$) below Component 4 ($\text{BS} = 1$); and Component 4 ($\text{BS} = 2$) below Component 4 ($\text{BS} = 1$).


\begin{figure}[ht]
\centering
\includegraphics[width=5in]{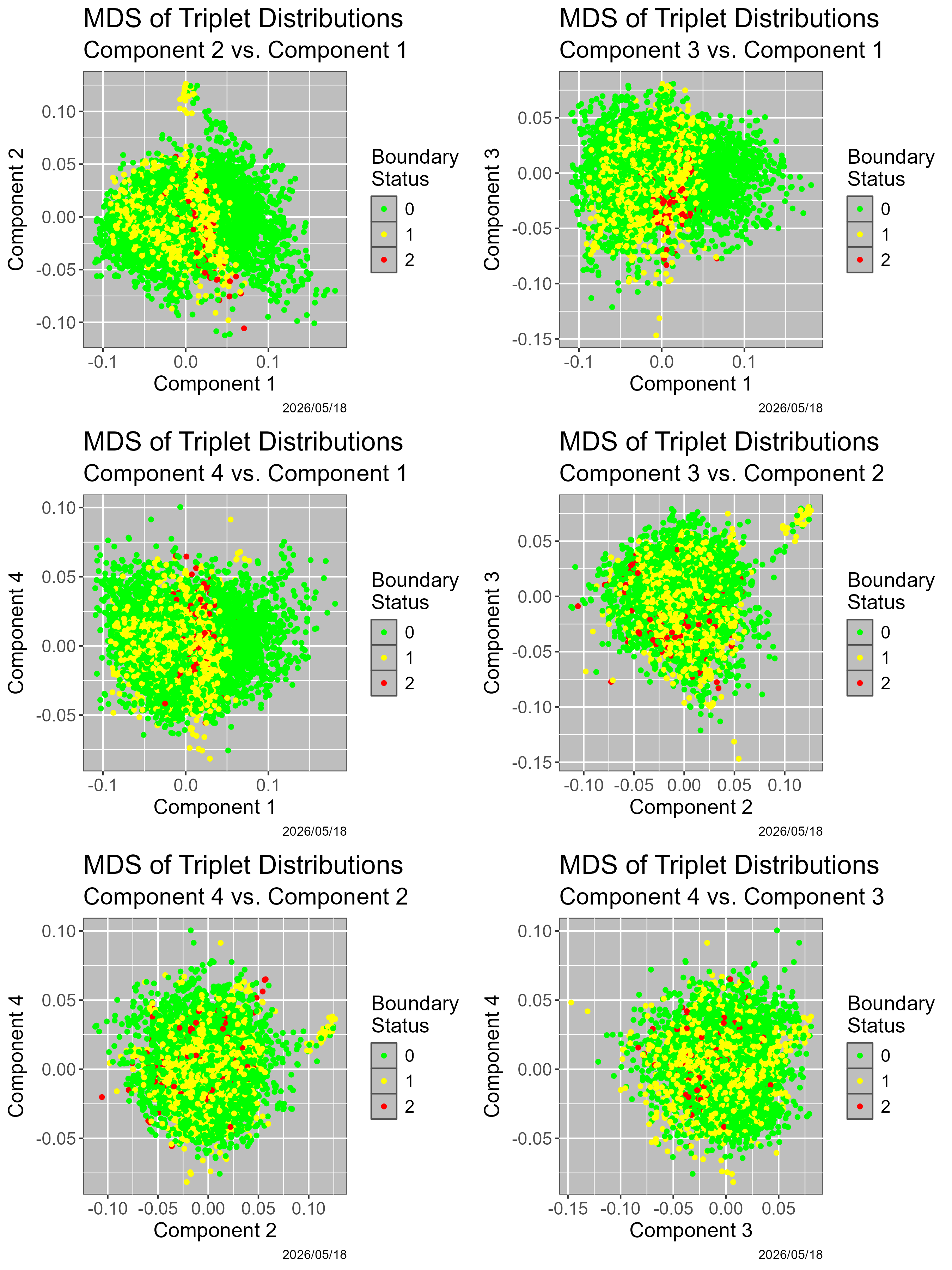}
\caption{Pairwise scatterplots of the first four MDS components of the triplet distributions, colored by Boundary Status.}
\label{fig.mds}
\end{figure}

\begin{figure}[ht]
\centering
\includegraphics[width=5in]{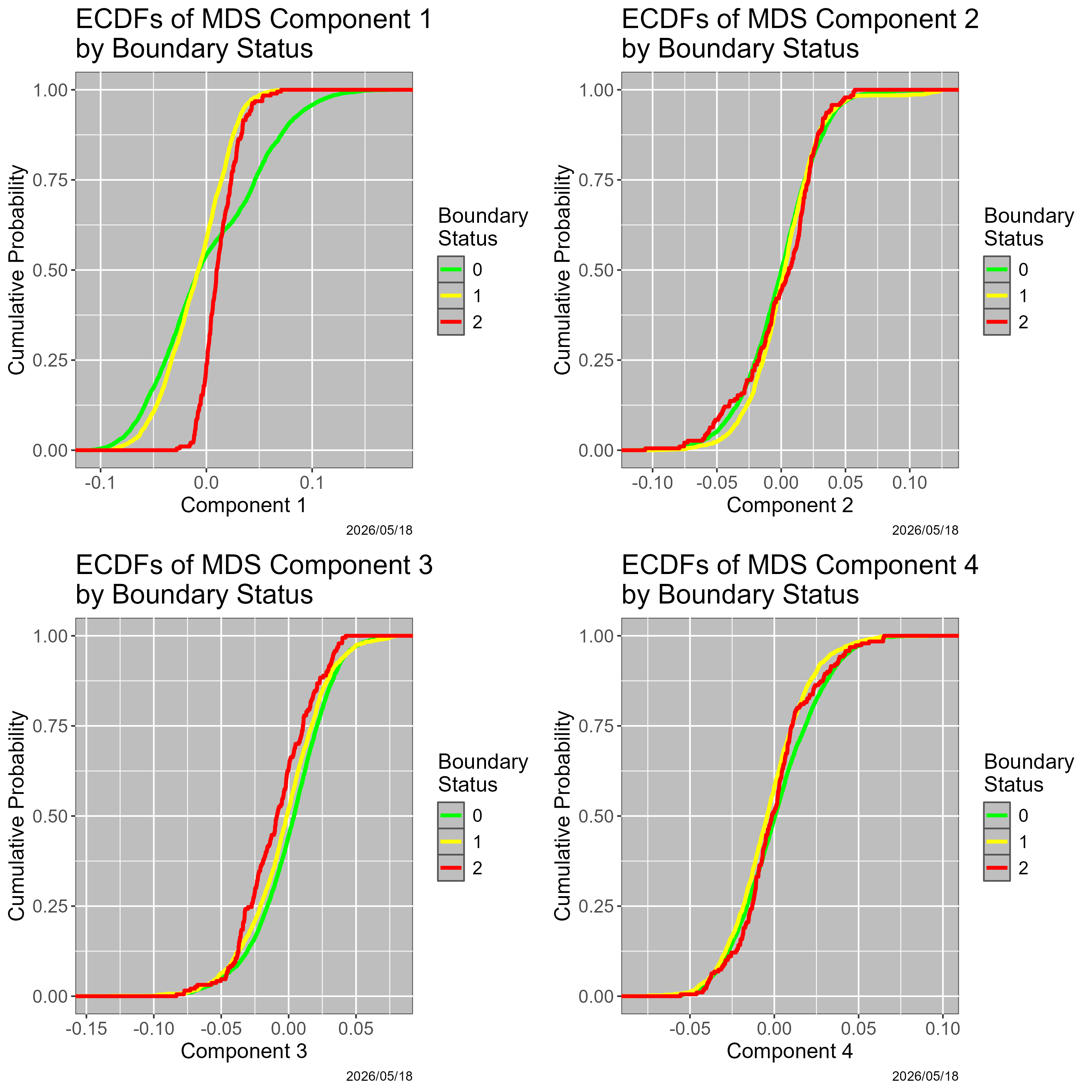}
\caption{ECDFs of the first four MDS components, colored by Boundary Status.}
\label{fig.mds-ecdf}
\end{figure}

A partition model with triplet distributions as predictors and $\text{BS}$ as response tells a complementary story. Without pruning, the model has 818 terminal nodes, with sizes ranging from 1 to 296, and predicts Boundary Status exactly. When the model is pruned with standard heuristics, there are 705 nodes ranging in size from 1 to 347, with the confusion matrix shown in shown in Table \ref{tab.partition-reads}. The correct classification rate is 98.76\%.

Many analysts would still be uncomfortable that this model is overfit, so in Table \ref{tab.pruning}, we show all possible pruned versions of the model, as generated by the \textsf{R} package \texttt{rpart}. At the top is the ``default model'' that predicts all reads to have Boundary Status 0. At the bottom is the unpruned, and perfect, model. Halving the incorrect classification rate from 30\% to 15\%, for instance, requires only 182 terminal nodes. Their sizes range from 3 to 445, with a mean size of 32 and median of 14. The first column in this table contains the complexity parameter, which quantifies the relationship between model complexity and model fit. More precisely, it is the improvement in model fit resulting from making additional splits. The table contains only values at which the structure of the optimal tree changes.


\begin{table}[ht]
\centering
\caption{Confusion matrix for the pruned partition model of Boundary Status as a function of triplet distribution, for the 5869 reads. The correct classification rate is 98.76\%.}
\label{tab.partition-reads}

\vspace{.1in}
\begin{tabular}{r|rrr|r}
\hline
& \multicolumn{3}{c|}{Predicted} & \\
Actual & 0 & 1 & 2 |& Sum \\
\hline
0 & 4061 & 17 & 0 & 4078 \\
1 & 49 & 1552 & 0 & 1601 \\
2 & 4 & 3 & 183 & 190 \\
\hline
Sum & 4114 & 1572 & 183 & 5869 \\
\hline
\end{tabular}
\end{table}

\begin{table}[ht]
\centering
\caption{For all possible pruned partition models, the number of terminal nodes and correct classification rate. The principal pruned version is indicated by ***, and the ``halve the error rate'' model by +++.}
\label{tab.pruning}

\vspace{.1in}
\begin{scriptsize}
\begin{tabular}{lrr}
  \hline
Complexity Parameter & Terminal Nodes & Correct Classification Rate \\
  \hline
  0.004467 & 1 & 69.483728 \\
  0.003908 & 35 & 74.646447 \\
  0.003350 & 41 & 75.413188 \\
  0.003210 & 42 & 75.515420 \\
  0.002792 & 49 & 76.299199 \\
  0.002606 & 61 & 77.321520 \\
  0.002513 & 65 & 77.645255 \\
  0.002233 & 74 & 78.394957 \\
  0.002171 & 110 & 80.882604 \\
  0.002122 & 123 & 81.802692 \\
  0.002047 & 128 & 82.126427 \\
  0.001954 & 131 & 82.313852 \\
  0.001861 & 140 & 82.859090 \\
  0.001675 & 143 & 83.029477 \\
  0.001535 +++  & 182 & 85.244505 \\
  0.001489 & 189 & 85.619356 \\
  0.001396 & 204 & 86.300903 \\
  0.001303 & 214 & 86.726870 \\
  0.001196 & 223 & 87.084682 \\
  0.001117 & 232 & 87.425456 \\
  0.000977 & 357 & 91.906628 \\
  0.000931 & 364 & 92.128131 \\
  0.000838 & 375 & 92.468904 \\
  0.000782 & 410 & 93.388993 \\
  0.000744 & 420 & 93.678651 \\
  0.000698 & 441 & 94.223888 \\
  0.000558 & 445 & 94.309082 \\
  0.000465 *** & 705 & 98.756177 \\
  0.000419 & 712 & 98.858409 \\
  0.000372 & 724 & 99.028795 \\
  0.000279 & 773 & 99.625149 \\
  0.000186 & 815 & 99.982961 \\
  0.000000 & 818 & 100.000000 \\
   \hline
\end{tabular}
\end{scriptsize}
\end{table}

\subsection{Is the Boundary Connected?}\label{subsec.connected}
It is natural to ask whether the boundary is connected, in the sense that for any two points on it, there is a path connecting them that consists entirely of boundary points. It is clear that a definitive answer is not feasible computationally. Our evidence to date is that, at least in the vicinity of the training data, the boundary is connected. Of multiple attempts to connect boundary points using a greedy algorithm and not leaving the boundary, none failed.

The algorithm appears in Algorithm \ref{alg.boundary}. To construct a path from $R_o \in \mathcal{B}$ to $R_d \in \mathcal{B}$ ($o$ is for origin, $d$ is for destination) that does not leave $\mathcal{B}$, simply change the classifier decision at each step, which ensures remaining within $\mathcal{B}$, and reduce the distance to $R_d$. If this is not possible, remain the same distance from $R_d$. And if this is not possible, stop and return an error. The latter was never necessary, among nearly 100 cases that varied with respect to read source, classifier decision, and correctness of the decisions for $R_o$ and $R_d$.

\begin{algorithm}
\caption{Algorithm for connecting boundary points without leaving the boundary.}
\label{alg.boundary}
\begin{algorithmic}[1] 
\Require Bayes classifier $C$ and boundary points $R_o, R_d \in \mathcal{B}$
\Ensure $BP(R_o, R_d)$
\State Path to be constructed: $BP = (R_o)$
\State Current point on path: $CP = R_o$
\State Current distance to $R_d$: $CD = H(CP, R_d)$, where $H$ = Hamming distance
\While{$CP \neq R_d$}
    \State Let $N(CP)$ be the set of all neighbors of $CP$
    \State $X \leftarrow \left\{R \in \mathcal{N}(CP): R \notin BP, C(R) \neq C(CP) \right\}$
    \State $X_1 \leftarrow \left\{R \in R: H(R^*, R_d) < H(CP, R_d) \right\}$
    \If{ $X_1 \neq \emptyset$ }
        \State Select $R^*$ randomly from $X_1$
        \State $BP \leftarrow (BP, R^*)$
        \State $CP \leftarrow R^*$
    \ElsIf{ $X  \neq \emptyset$ }
         \State Select $R^*$ randomly from $X$
         \State $BP \leftarrow (BP, R^*)$
        \State $CP \leftarrow R^*$
    \Else
         \State \Return Error
    \EndIf
\EndWhile
\State \Return $BP$
\end{algorithmic}
\end{algorithm}

\subsection{Scientific Generalizability}\label{subsec.generality}
There are, of course, questions of scientific generalizability: are the properties observed somehow specific to our read dataset, or descriptive of the classifier $C$ more generally? And do they apply to other classifiers? Do they extend to other forms of data?

\subsubsection{Other Datasets}\label{subsubsec.other-datasets}
To address the first question, we ran the Bayes classifier on a number of additional datasets, which are described in Table \ref{tab.other-datasets-description}. Results appear in Table \ref{tab.other-datasets-results}. For completeness, the training dataset itself is in the first row. All genomes were downloaded from \NCBI.

These alternative datasets fall into four classes:
\begin{description}
\item[Validation:]
This dataset is effectively a clone of the training dataset, and has been used in
\cite{amost2024}, \cite{classifier-tdq2026}, and elsewhere to to evaluate model performance, which should (and does) match that for the training data.
\item[Random sequences,]
to possibly reach regions of $\mathcal{I}$ in which ``real'' DNA sequences are sparse.
\item[Local behavior,]
in Distance2, examining near-neighbors of Validation.
\item[Other genomes,]
which include other viruses, bacteria and the human genome.
\end{description}
For each, there are both one dataset of the nature just described and a second constructed by randomly sampling Hamming paths between 400 randomly selected pairs, which represent a more localized sampling of $\mathcal{I}$.

Most important, all of the alternative datasets can be regarded as methods for sampling randomly from $\mathcal{I}$.

\begin{table}[ht]
\centering
\caption{Short descriptions of alternative datasets.}
\label{tab.other-datasets-description}

\vspace{.1in}
\begin{tabular}{l|p{3in}|r}
\hline
Name & Description & Size
\\
\hline
Training & Used throughout & 5869
\\
\hline
Validation & 2000 Mason simulator-generated length-101 reads each from Adeno, COVID and SARS genomes & 6000
\\
Random & Random sequences of length 101 & 10000
\\
Random2 & Random sequences of length 100 & 25000
\\
Distance2 & Two randomly generated Hamming distance 2 neighbors of each elements of Validation & 12000
\\
3Viruses & Mason simulator-generated reads from norovirus, human and H5N1 influenza virus & 1713
\\
Ecoli & Length 101 Mason simnulator-generated reads from an \emph{Escherichia coli} genome & 12000
\\
Pg & Length 101 Mason simnulator-generated reads from an \emph{Porphyromonas gingivalis} genome & 25000
\\
Hsc1 & Length 101 Mason simnulator-generated reads from an \emph{Home sapiens} chromosome 1 & 25000
\\
\hline
TrainingPaths & All elements of a random sample of Hamming paths between 400 randomly selected pairs of Training & 30188
\\
ValidationPaths & All elements of a random sample of Hamming paths between 400 randomly selected pairs of Validation & 30452
\\
RandomPaths & All elements of a random sample of Hamming paths between 400 randomly selected pairs of Random & 30697 \\
Random2Paths & All elements of a random sample of Hamming paths between 400 randomly selected pairs of Random2 &  30590 \\
Distance2Paths & All elements of a random sample of Hamming paths between 400 randomly selected pairs of Distance2 & 30343 \\
3VirusesPaths & All elements of a random sample of Hamming paths between 400 randomly selected pairs of 3Viruses & 30440  \\
EcoliPaths & All elements of a random sample of Hamming paths between 400 randomly selected pairs of Ecoli & 30713 \\
PgPaths & All elements of a random sample of Hamming paths between 400 randomly selected pairs of Pg & 30776 \\
Hsc1Paths & All elements of a random sample of Hamming paths between 400 randomly selected pairs of Hsc1& 30497 \\
\hline
\end{tabular}
\end{table}

Boundary Status distributions for each dataset in Table \ref{tab.other-datasets-description} appear in Table \ref{tab.other-datasets-results}. The most striking feature is that only for the Ecoli datasets does less 20\% of the dataset lie on the boundary, and then only slightly. We interpret this as further evidence supporting our claim that the boundary is large. Moreover, for some datasets, one-half of the points lie on the boundary.

The results for Distance2, and to a lesser extent 3Viruses, show an important insight that at first may seem paradoxical. The boundary is denser in the vicinity of the Training and Validation, and sparser for Random, Random2, Ecoli, Pg and Hsc1. The reason for this, as discussed in detail and from a slightly different perspective in \cite{classifier-tdq2026}, is that in regions distant from the training data  the classifier degenerates. In this case, as demonstrated by the ``Frac(Adeno)'' column in Table \ref{tab.other-datasets-results}, nearly everything that is distant from the training data is classified as Adeno. Hence, there is no paradox: near the training data, we want high discrimination, which means a larger boundary. Also, therefore, Boundary Status is usable as a diagnostic indicating when a classifier is being used on data that differ from what was ``expected.'' For the Bayes classifier, ``what is expected'' is incorporated in the prior and the likelihood functions \citep{markovstructure-2021}. For other classifiers, the training data represent ``what is expected.'' Finally, of course, one must temper interpreting high Neigbhor Similarity as confidence. This works only when the analysis data are not too far from ``expected.''

For full disclosure, we note that Hsc1 does not differ from the training data nearly as much as do Random, Random2, Ecoli and Pg. We are not able to explain this, nor to hypothesize a plausible explanation for it.

Finally, we note that the material here complements that in Section \ref{subsec.results-reads} in that they are alternative methods to explore the boundary.

\begin{table}[ht]
\centering
\caption{Boundary Status Distributions for Bayes classifier applied to alternative datasets. Columns Boundary Status are percentages. The column Frac(Adeno) is fraction of predictions that are Adenovirus.}
\label{tab.other-datasets-results}

\vspace{.1in}
\begin{tabular}{l|rrr|r}
\hline
 & \multicolumn{3}{|c|}{BoundaryStatus} & Frac(Adeno)
\\
Dataset & 0 & 1 & 2
\\
\hline
Training & 0.695 & 0.273 & 0.032 & 0.329
\\
Validation & 0.687 & 0.282 & 0.03 & 0.324
\\
Random & 0.792 & 0.196 & 0.012 & 0.912
\\
Random2 & 0.789 & 0.201 & 0.01 & 0.911
\\
Distance2 & 0.672 & 0.293 & 0.034 & 0.334
\\
3Viruses & 0.48 & 0.489 & 0.031 & 0.472
\\
Ecoli & 0.877 & 0.107 & 0.017 & 0.928
\\
Pg &  0.76 & 0.218 & 0.023 & 0.856
\\
Hsc1 & 0.554 & 0.381 & 0.064 & 0.456
\\
\hline
TrainingPaths & 0.575 & 0.352 & 0.073 & 0.348
\\
ValidationPaths & 0.594 & 0.346 &  0.06 & 0.384
\\
RandomPaths & 0.798 & 0.194 & 0.009 & 0.916
\\
Random2Paths & 0.792 & 0.196 & 0.012 & 0.91
\\
Distance2Paths & 0.589 & 0.336 & 0.075 & 0.4
\\
3VirusesPaths & 0.49 & 0.465 & 0.044 & 0.556
\\
EcoliPaths & 0.828 & 0.152 & 0.021 & 0.916
\\
PgPaths & 0.752 & 0.224 & 0.024 & 0.868
\\
Hsc1Paths & 0.514 & 0.406 &  0.08 & 0.487
\\
\hline
\end{tabular}
\end{table}


\subsubsection{Other Classifiers}\label{subsubsec.other-classfiers}
Full results will be reported elsewhere \citep{classifier-tdq2026}, but initial experiments show similar behavior for other classifiers applied to the same dataset, including partition models, random forests and neural networks \citep{amost2024}. Table \ref{tab.comparison4}, which also appears there, shows that performance of four classifiers on the training data, with respect to correct classification rate, size of the boundary and mean Neighbor Similarity.

\begin{table}[ht]
\centering
\caption{Comparison of four classifiers. CCR = correct classification rate; $|\mathcal{B}|$ = size of boundary;  $\overline{NS}$ = mean Neighbor Similarity.}
\label{tab.comparison4}

\vspace{.1in}
\begin{tabular}{l|rrr}
\hline
Classifier & CCR & $|\mathcal{B}|$ & $\overline{NS}$
\\
\hline
Bayes & .8155 & 1791 & .9292
\\
Partition model & .8075 & 4484 & .8393
\\
Random forest & 1.0000 & 2173 & .9897
\\
Neural net & .7657 & 2154 & .9022
\\
\hline
\end{tabular}
\end{table}

\subsubsection{Other Forms of Data}\label{subsubsec.other-dataforms}
The essential characteristics of our framework are the graph structure of $\mathcal{I}$, which requires discreteness, and discreteness of $\mathcal{O}$. To illustrate, $\mathcal{I}$ could be the space of all $1024 \times 1024$ RGB images with 8-bit color depth. Therefore, no one should be surprised that there are images that look the same but are classified differently. Indeed, because there is a boundary, there are image pairs that differ by one color level in one pixel that are classified differently. Of course in this setting, computational issues preclude anything resembling the analyses here.

\section{Discussion}\label{sec.discussion}
The extent to which lying on or near the boundary $\mathcal{B}$ can or should affect either classifier output or decisions based on it is contextual. The first issue is whether to check Boundary Status for a given query. If one cares only about whether the query lies on the boundary, the process can be terminated once a neighbor with a different result is found. However, given the size of the boundary, this strategy is almost surely too blunt. In many cases, both scientific and broader impacts merit a quantified measure such as Neighbor Similarity. Although in our read classification case study, the consequences of misclassification are arguably not major, for classifiers such as \texttt{SeqScreen} \citep{seqscreen2022}, which is designed to identify pathogenic sequences, the consequences of incorrect decisions may be dramatic, and the risks of accepting high-uncertainty classifier results without further investigation may be too high in such contexts. Not knowing the level of uncertainty is equally problematic.

Furthermore, in the face of rampant concern about data quality in DNA databases \citep{critical-assessment-2021, langdon-2014, steinegger-2020}, and given that changing a data point to a neighbor, as done in \citep{dqdegradation-2021} to measure data quality, is seemingly the most innocuous data quality problem, there is reason for concern. This is so at least with respect to the accuracy dimension of data quality, which is but one of many dimensions \citep{dq-statmeth06}. Therefore, it seems prudent to care about neighbors. The issue is discussed at length in \cite{classifier-tdq2026}.

In Section \ref{sec.results}, we showed that that Neighbor Similarity relates to both inherent measures of uncertainty for our Bayes classifier. From an operational perspective, Figure \ref{fig.roc} is equally important. It contains \ROC\ curves for two decision rules, one based on accepting decisions for which $\text{MAP}$ exceeds a threshold, which is encoded by color, the other for the decision rule  based on accepting decisions for which $\text{NS}$ exceeds a threshold. The two \ROC\ curves are essentially identical. For the Bayes classifier, $\text{NS}$ is as effective as $\text{MAP}$ for making principled decisions. 

Although this point is addressed elsewhere \citep{amost2024, classifier-tdq2026}, we emphasize that $\text{NS}$ is implementable for all classifiers.

\begin{figure}[ht]
\centering
\includegraphics[width = 2in]{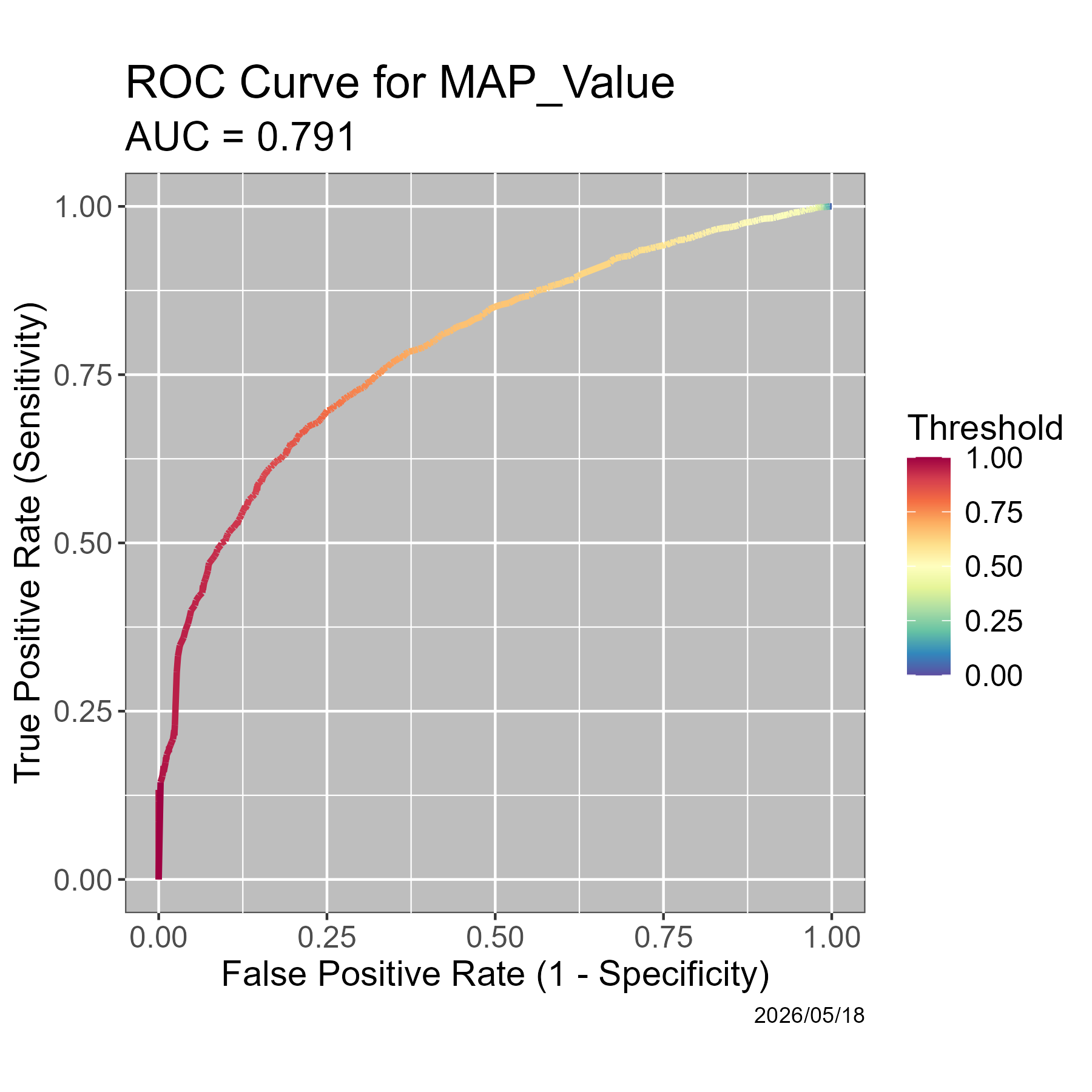} \hspace{.25in} \includegraphics[width = 2in]{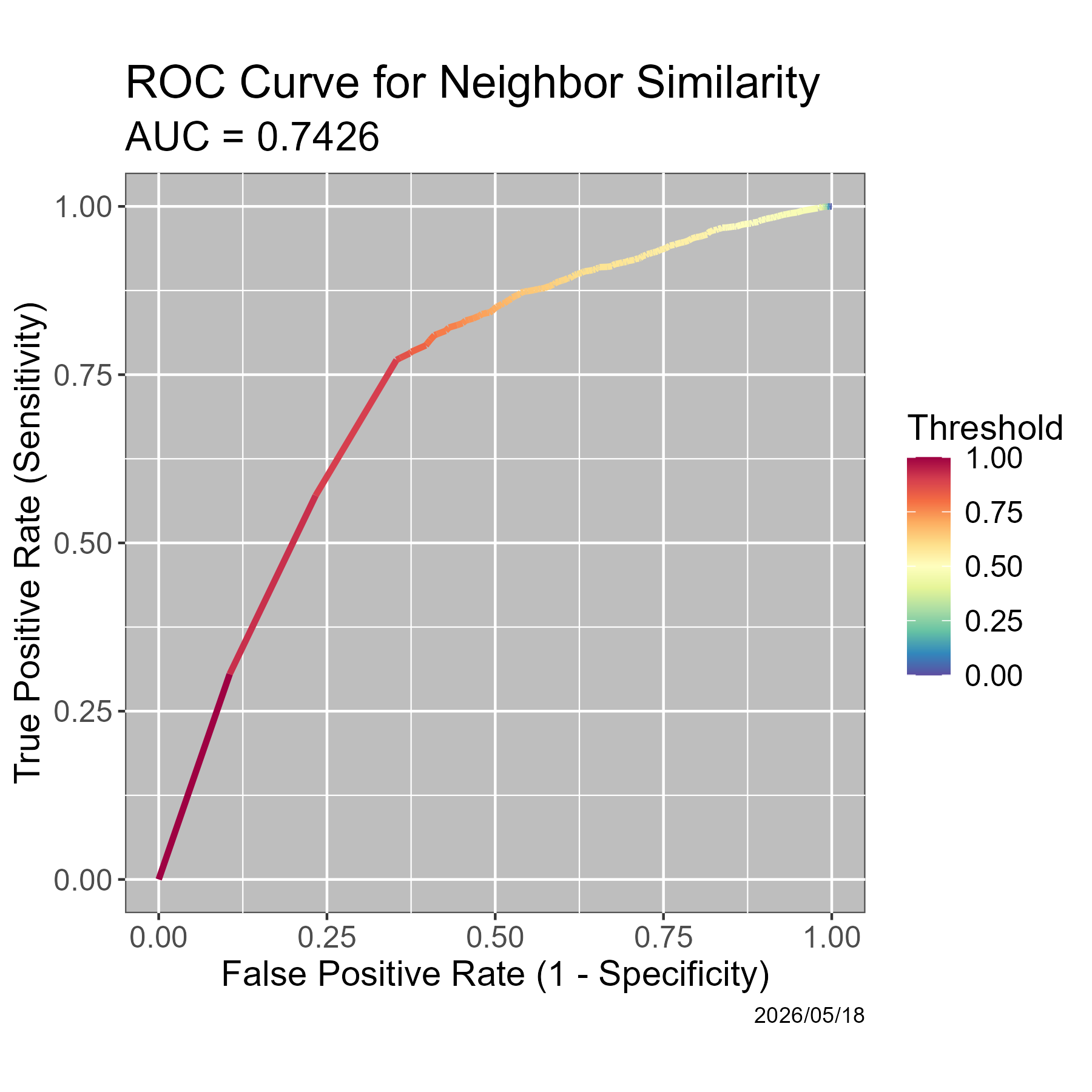}
\caption{\emph{Left:} ROC curve for the decision rule based on thresholded values of $\text{MAP}$. \emph{Right:} ROC curve for the decision rule based on thresholded value of $\text{NS}$. In each panel, the $x$-axis is the False Positive Rate (1 - Specificity) and the $y$-axis is the True Positive Rate (Sensitivity).}
\label{fig.roc}
\end{figure}

\section{Conclusion}\label{sec.conclusion}
The three principal contributions of this paper are the concept of the boundary, surrogate measures of uncertainty for classifiers and methods for investigating boundaries. In particular, we propose Neighbor Similarity as a measure of uncertainty for classifiers lacking inherent measures of uncertainty. Despite the computational burden, the potential to avoid incorrect decisions is significant.

As always, at least as many questions are raised as answered. One is whether Neighbor Similarity is viable in the context of production-grade classifiers such as \texttt{SeqScreen}. A central issue may be whether the information gained justifies the computational cost. A (still short) DNA sequence of length 10,000 has more than 40,000 neighbors, and running SeqScreen on all of them may be prohibitive. We showed the potential of sampling to estimate Neighbor Similarity from a subset of neighbors; investigating demonstrably effective strategies to do so is a priority.

Underlying everything is a scientific and statistical issue not addressed in this paper and clearly needing more attention. Is a low value of $\text{NS}(R)$ a statement about the read or about the classifier? Or, if present, about the training data? As long there is a range of observed values of $\text{NS}$, and especially if there are values near 1, then low values are plausibly statements about reads. Stronger evidence that a read is problematic would be that it is on the boundary for multiple classifiers, which initial evidence shows to be true \citep{amost2024}. On the other hand, uniformly small values of $\text{NS}$ seem to suggest problems with the classifier (or, if present, the training data). But then, what if the problem is just innately hard?

\section*{Acknowledgements}
This research was supported in part by NIH grant 5R01AI100947--06, ``Algorithms and Software for the Assembly of Metagenomic Data,'' to the University of Maryland College Park (Mihai Pop, PI). We thank Professor Pop for numerous insightful discussions.

\newpage\clearpage
\begin{appendices}
\section{Triplet Distributions of the Three Genomes}\label{app.TD}
\begin{table}[ht]
\centering
\caption{Triplet distributions for the adenovirus, COVID and SARS genomes used in this paper.}

\vspace{.1in}
\begin{scriptsize}
\begin{tabular}{lrrr||lrrr}
  \hline
Triplet & Adeno & COVID & SARS & Triplet & Adeno & COVID & SARS \\
  \hline
  AAA & 0.031826 & 0.026367 & 0.025581 & GAA & 0.017496 & 0.012899 & 0.015261 \\
  AAC & 0.020016 & 0.010660 & 0.018051 & GAC & 0.012220 & 0.005982 & 0.012337 \\
  AAG & 0.018463 & 0.016776 & 0.018891 & GAG & 0.013832 & 0.007252 & 0.013278 \\
  AAT & 0.018551 & 0.030176 & 0.021917 & GAT & 0.011927 & 0.021621 & 0.015597 \\
  ACA & 0.020016 & 0.012699 & 0.026219 & GCA & 0.017847 & 0.008221 & 0.014421 \\
  ACC & 0.015327 & 0.006851 & 0.013076 & GCC & 0.015063 & 0.004278 & 0.007866 \\
  ACG & 0.010316 & 0.003709 & 0.005210 & GCG & 0.015327 & 0.002373 & 0.004975 \\
  ACT & 0.016265 & 0.016609 & 0.022018 & GCT & 0.016499 & 0.013868 & 0.020908 \\
  AGA & 0.014213 & 0.016208 & 0.018085 & GGA & 0.016704 & 0.005514 & 0.011698 \\
  AGC & 0.017701 & 0.007519 & 0.011765 & GGC & 0.014243 & 0.005247 & 0.009849 \\
  AGG & 0.015591 & 0.008421 & 0.013984 & GGG & 0.012279 & 0.003409 & 0.004975 \\
  AGT & 0.014711 & 0.019015 & 0.015059 & GGT & 0.013041 & 0.018346 & 0.013916 \\
  ATA & 0.012836 & 0.024429 & 0.013345 & GTA & 0.013334 & 0.020151 & 0.014723 \\
  ATC & 0.011224 & 0.010694 & 0.011294 & GTC & 0.010228 & 0.007419 & 0.009984 \\
  ATG & 0.017378 & 0.029475 & 0.026085 & GTG & 0.014067 & 0.016308 & 0.018488 \\
  ATT & 0.018990 & 0.038765 & 0.024438 & GTT & 0.016294 & 0.037562 & 0.019698 \\
  CAA & 0.020573 & 0.012565 & 0.024471 & TAA & 0.018961 & 0.032148 & 0.019160 \\
  CAC & 0.014008 & 0.006015 & 0.016202 & TAC & 0.015679 & 0.017244 & 0.019933 \\
  CAG & 0.019342 & 0.008956 & 0.014891 & TAG & 0.010579 & 0.018179 & 0.011832 \\
  CAT & 0.017115 & 0.012030 & 0.018589 & TAT & 0.012836 & 0.039533 & 0.019026 \\
  CCA & 0.019400 & 0.006149 & 0.013311 & TCA & 0.013774 & 0.012498 & 0.020202 \\
  CCC & 0.014067 & 0.002573 & 0.004773 & TCC & 0.013744 & 0.006115 & 0.007059 \\
  CCG & 0.010257 & 0.001604 & 0.003059 & TCG & 0.008059 & 0.003676 & 0.005849 \\
  CCT & 0.014506 & 0.009491 & 0.011631 & TCT & 0.014301 & 0.019416 & 0.019093 \\
  CGA & 0.009055 & 0.002406 & 0.004672 & TGA & 0.015503 & 0.023593 & 0.022018 \\
  CGC & 0.015503 & 0.001771 & 0.004471 & TGC & 0.017290 & 0.014203 & 0.022085 \\
  CGG & 0.010257 & 0.001571 & 0.002756 & TGG & 0.018170 & 0.019115 & 0.018723 \\
  CGT & 0.009143 & 0.005614 & 0.007194 & TGT & 0.017027 & 0.038464 & 0.026724 \\
  CTA & 0.013012 & 0.017544 & 0.018723 & TTA & 0.018873 & 0.044981 & 0.023160 \\
  CTC & 0.011459 & 0.007085 & 0.012000 & TTC & 0.016968 & 0.016508 & 0.018925 \\
  CTG & 0.017525 & 0.014203 & 0.018891 & TTG & 0.019019 & 0.035390 & 0.026085 \\
  CTT & 0.019576 & 0.020552 & 0.024034 & TTT & 0.030595 & 0.059985 & 0.027463 \\
   \hline
\end{tabular}
\end{scriptsize}
\end{table}




\end{appendices}


\end{document}